\documentclass[10pt,journal,compsoc]{IEEEtran}
\ifCLASSOPTIONcompsoc
  \usepackage[nocompress]{cite}
\else
  \usepackage{cite}
\fi
\ifCLASSINFOpdf
\else
\fi

\usepackage{amsmath,amsfonts}
\usepackage{algorithmic}
\usepackage{algorithm}
\usepackage{array}
\usepackage{textcomp}
\usepackage{stfloats}
\usepackage{url}

\usepackage{verbatim}
\usepackage{graphicx}
\usepackage{cite}
\usepackage{caption}
\usepackage{multirow}
\usepackage{color}
\usepackage{subcaption}

\newcommand\equalcontribution{\thanks{S. Cao and W. Luo are the co-first authors. W. Zhang and L. Ma are the corresponding authors.}}

\newcommand{\eg}[0]{\textit{e.g.}}
\newcommand{\etc}[0]{\textit{etc}}
\newcommand{\ie}[0]{\textit{i.e.}}

\usepackage{soul}

\begin{document}
\title{A Circular Window-based Cascade Transformer for Online Action Detection}

\author{Shuqiang Cao$^{\dag}$, Weixin Luo$^{\dag}$\equalcontribution, Bairui Wang, Wei Zhang$^{\S}$, Lin Ma$^{\S}$
% <-this % stops a space
\IEEEcompsocitemizethanks{\IEEEcompsocthanksitem S. Cao and W. Zhang are with the School of Control Science and
Engineering, Shandong University, China. E-mail: sqiangcao@mail.sdu.edu.cn, davidzhang@sdu.edu.cn.
\IEEEcompsocthanksitem W. Luo, B. Wang, and L. Ma are with Meituan, China. E-mail: luowx@shanghaitech.edu.cn, \{bairuiwong, forest.linma\}@gmail.com.
}% <-this % stops an unwanted space
}

\IEEEtitleabstractindextext{
\begin{center}\setcounter{figure}{0}
    \includegraphics[width=17cm]{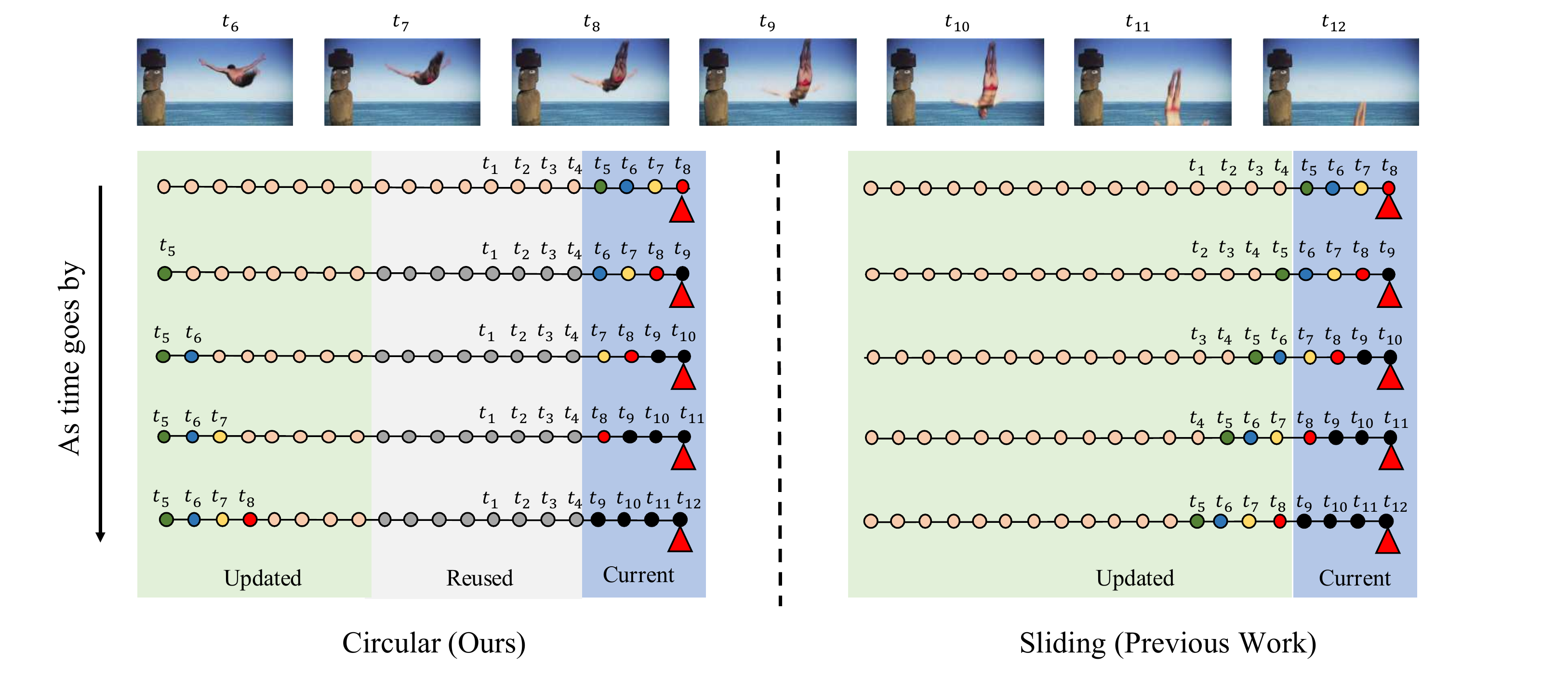}
    \captionof{figure}{\label{fig:teaser}Comparison of our key novelty and previous work for online action detection, which aims to predict action for the current frame (red triangle) at each step. As time goes by, our queue (left figure) pushes the latest history frame (\eg, $t_5$) to the updated window with a circular manner but reuses the intermediate. In contrast, the sliding queue (right figure) used in previous work rolls all the frames temporally, resulting in substantial computations.}
\end{center}

\begin{abstract}
Online action detection aims at the accurate action prediction of the current frame based on long historical observations. Meanwhile, it demands real-time inference on online streaming videos. In this paper, we advocate a novel and efficient principle for online action detection. It merely updates the latest and oldest historical representations in one window but reuses the intermediate ones, which have been already computed. Based on this principle, we introduce a window-based cascade Transformer with a circular historical queue, where it conducts multi-stage attentions and cascade refinement on each window. We also explore the association between online action detection and its counterpart offline action segmentation as an auxiliary task. We find that such an extra supervision helps discriminative history clustering and acts as feature augmentation for better training the classifier and cascade refinement. Our proposed method achieves the state-of-the-art performances on three challenging datasets THUMOS'14, TVSeries, and HDD. Codes will be available after acceptance.
\end{abstract}
}
\maketitle
\IEEEdisplaynontitleabstractindextext

% \fi
\IEEEpeerreviewmaketitle

\IEEEraisesectionheading{\section{Introduction}\label{sec:introduction}}

% \IEEEPARstart{R}{ecent} transformer-based methods significantly boost the performance, which separately design a long-term feature bank for historical information and a short-term window for recent contextual information. Then real-time history updates on stream video inputs will be realized by matrix decomposition on attention operations. Shown in Figure~\ref{fig:update_mechanism},
%介绍OAD任务定义，并引出历史信息的重要性。

% Though a line of recent work~\cite{xu2019temporal,eun2020learning,geest2016online,gao2017red} has achieved great progress, such a promising task still confront with a significant challenge as the future frames are unavailable when classifying the observed frame at the moment. The observed history frames, however, are recorded so that they play as supplementary to the current frame. Thus, how to tackle the observed history and predict the unobserved future has being become one of primary research aspects in this task, which motivates us in this paper.  
%已有OAD怎么处理历史信息，有什么缺点。
Recent work has employed Recurrent Neural Networks~(RNN), \eg, Long Short-Term Memory (LSTM)~\cite{hochreiter1997long} or Gated Recurrent Unit (GRU)~\cite{cho2014learning} to model the temporal dependency. However, the gate mechanism they deploy is much more likely to forget the oldest information that may still be valuable for the current classification as time goes by~\cite{wu2019long}. Many recent methods~\cite{xu2021long,chen2022gatehub} introduce Transformer architecture to avoid this problem by attending to a long-term historical feature bank all the time. The powerful Transformer brings significant improvement in performance for online action detection (OAD). However, its quadratic complexity becomes an obstacle when tackling long sequence inputs of OAD. To alleviate this problem, one of the state-of-the-art methods LSTR~\cite{xu2021long} leverages a cross-attention layer~\cite{jaegle2021perceiver} to compress the long historical feature bank to a limited number of latent representations. Since there is only one new frame to be updated at each time step, it also introduces a matrix decomposition on the initial attention layer to avoid redundant computation during the temporal movement of stream videos. Though it has promising performance, the naive cross-attention design to compress long-term history may not be optimal, because it misses so many advanced architectures such as hierarchical stages~\cite{simonyan2014very,he2016deep}, cascade~\cite{farha2019ms}, and feature pyramid~\cite{lin2017feature} of which effectiveness has been verified for video understanding.

To support the real-time online inference, it is desirable to design a highly efficient updating strategy for streaming inputs. Because data obeys a First-In-First-Out (FIFO) mechanism, previous works~\cite{xu2021long}~\cite{chen2022gatehub} naturally adopt a sliding updating strategy on the FIFO queue, as shown in Figure~\ref{fig:teaser}. However, we cannot bear the large computation costs for updating the entire long history, though they accelerate online inference by storing the intermediate results of the attention operation at the first stage. In addition to the new coming and dropped frames, we \textbf{observe} that the intermediate frames in the long history are unchanged at each step. Thus, we can reuse the previous results of these intermediate frames at this moment until they are abandoned. To update the representation of the new coming frame, we concatenate it with the oldest history frames within a circular window with a limited size. Such a simple strategy allows us to perform only one window updating with a deep architecture, \eg, hierarchical stages and cascade but low computation costs due to the limited number of tokens in one window.  

We also \textbf{observe} the relationship between the OAD and Offline Action Segmentation~\cite{farha2019ms} (OAS). The biggest difference between them is the accessibility to the future frames when performing per-frame classification. For the OAD task, most of history frames can access their ``future`` observations. Thus, the first association between these two tasks is to perform OAS on the history. Such an auxiliary task enforces the compressed historical representatives to perform per-frame classification, leading to memorization of all the historical cues. Second, these two tasks share the same objective function, \ie, per-frame action classification. Therefore, we try to associate them by sharing the modules with the same functions, such as the action classifier and cascade refinement, where the OAS task can be treated as feature augmentation for better training.

%具体来说我们怎么做，循环窗口，MTSM，cascade，多任务学习，
Based on these two \textbf{observations}, this paper proposes a novel Circular Window-based Cascade Transformer (CWC-Trans) for effective and efficient online action detection. Specifically, we split the entire feature bank into two components, \ie, a history feature bank with multiple windows and a current feature window including the incoming frame. The long history bank will be abstracted to a limited number of clusters attended by the current window for the incoming frame classification. For each historical window, we deploy a hierarchical network with multiple stages. To refine the coarse predictions, we introduce a cascade module that is also conducted within windows so that it is able to update one window only at each time step. Also, it is built on the coarse predicted probabilities rather than the features. Thus, the shallow refinement module resides in a low-dimension space, and it does not bring high extra computation costs.  

We conduct extensive experiments on three public datasets THUMOS'14~\cite{HaroonIdrees2017TheTC}, TVseries~\cite{geest2016online}, and HDD~\cite{ramanishka2018CVPR}. Our proposed CWC-Trans achieves state-of-the-art performance on these three datasets. For example, CWC-Trans significantly outperforms LSTR~\cite{xu2021long} and GateHUB\cite{chen2022gatehub} on THUMOS'14, while it achieves promising results on TVseries and HDD. It also realizes 80.7 FPS for model inference, which supports our motivation behind this paper, \ie, to realize an effective and efficient OAD.

We further summarize four contributions we make as follows: i) we advocate a circular window partition strategy for real-time stream video processing, which only needs to conduct updates in one window for increment. Such a novel strategy allows us to develop elaborate models like hierarchical stages and cascade refinement without introducing high computation costs during online inference; ii) we introduce a novel multi-head token slim module for hierarchical modeling, which dynamically selects the crucial tokens via attention; iii) an auxiliary action segmentation task on historical feature bank is deployed to learn compact and discriminative historical clusters. The shared classifier on the main and auxiliary enhances its generalization ability. We claim that this is the first work to leverage the associations between these two tasks; iv) CWC-Trans achieves state-of-the-art performance on all the datasets, and we provide extensive experiments to verify each component proposed. 

The rest of the manuscript is organized as follows. In Section 2, we study related work, including multiple research topics such as online and offline action analysis, Transformer, \etc. Section 3 introduces our proposed CWC-Trans and demonstrates each module in detail. In Section 4, experiments on three publicly available datasets are conducted to validate the effectiveness of CWC-Trans. We conclude in Section 5.

\section{Related Works}

\subsection{Online Action Detection}
Online action detection focuses on untrimmed streaming videos with only current and historical information. This task is first introduced by De Geest ~\textit{et al.} \cite{geest2016online} who propose a new dataset TVseries for this task and compare several methods on it. 
%Afterwards, \textcolor{blue}{they propose a two-stream feedback network \cite{de2018modeling} to model the input features and temporal structure separately.}
Afterwards, they propose to model the appearance and the temporal structure of the input sequence~\cite{de2018modeling} using a two-stream paradigm~\cite{simonyan2014two}. 
Gao~\textit{et al.} \cite{gao2017red} design a Reinforced Encoder-Decoder(RED) model with a reinforcement reward function. RED incorporates action classification and anticipation in a single framework. Specifically, it takes history information as input and conducts the anticipation of the future frames after a few seconds. Besides, the reinforcement reward function encourages identifying actions as early as possible. 
%Xu~\textit{et al.} \cite{xu2019temporal} proposed a LSTM-based model, called 
TRN \cite{xu2019temporal} and OadTR \cite{wang2021oadtr} also utilize predicted future information by performing action anticipation to better recognize actions taking place in the present. 
%Specifically, TRN first predicts future actions as an anticipation task and then uses features for them and historical evidence together for detecting the current action. 
Since online action detection only accesses the frames in the current and past, recent work~\cite{eun2020learning,qu2020lap,xu2021long} attempt a more efficient way to aggregate these two features. For example, IDN~\cite{eun2020learning} replaces regular GRU~\cite{cho2014learning} units with a novel information discrimination unit to better aggregate information relevant to the present. 
LAP-Net~\cite{qu2020lap} proposes an adaptive future feature sampling strategy that dynamically determines the range of future contexts by estimating the current action process.
LSTR~\cite{xu2021long} employs a long-term and short-term memory mechanism to model frame sequences using Transformer. WOAD~\cite{gao2021woad} tackles this task with video-level labels using weak supervision. Colar~\cite{yang2022colar} designs an exemplar-consultation mechanism for online action detection, aggregating the historical and categorical exemplars. GateHUB~\cite{chen2022gatehub} develops a gated history unit to enhance the relevant information and suppress the noise in the history. 

% 对比一下LSTR和我们方法的差异; (已有的方法是怎样做的? 存在怎样的缺陷, 举个例子? 我们就有了动机，引出我们的Window-based 方法的优, 举个例子来说明?)；
These Transformer-based approaches~\cite{wang2021oadtr,xu2021long,chen2022gatehub} have achieved promising results on OAD tasks. However, because of the square complexity of self-attention with respect to the sequence length, it is unlikely to encode complex dependencies in long historical sequences by stacking a bulk of attention layers. For instance, OadTR~\cite{wang2021oadtr} samples a small set of the most recently observed frames so that three encoding layers can be used to represent the samples. %LSTR~\cite{xu2021long} introduces a long- and short-term memories mechanism to model the history information. 
LSTR~\cite{xu2021long} adopts Perceiver~\cite{jaegle2021perceiver} to compress long-term history, but the large-scale token reduction cannot keep all the critical information.
Besides, the compression relies on the data-agnostic queries, which does not consider the intrinsic properties of input data. 
We claim that our proposed single window updating can remedy these issues. It allows us to stack multiple attention layers due to the limited size of a window and to use long-term historical information at each time step since we reuse most of the historical representation already computed in the previous period.  
% Therefore, we design a \textbf{hierarchical attention} mechanism to progressively compress the history information in each separated window. Meanwhile, to ensure the efficiency of online streaming inference, the attention only performed on the updated window that embodies the latest and the oldest historical representations. 

\subsection{Offline Action Segmentation}
% MS-TCN
Action Segmentation aims at per-frame action classification in untrimmed videos. Earlier methods adopt a sliding window-based approach~\cite{MarcusRohrbach2012ADF,karaman2014fast} with non-maximum suppression to filter out the redundant windows. However, these methods cannot establish long-term temporal dependencies. To this end, the researchers use hidden Markov model(HMM)~\cite{HildeKuehne2015AnEG,KevinTang2012LearningLT}, RNN~\cite{HildeKuehne2020AHR,BharatSingh2016AMB}, and spatial-temporal convolution~\cite{ColinLea2016SegmentalSC} to model the long-range context. Temporal convolutional network (TCN) has been widely used in the field of speech synthesis~\cite{AaronvandenOord2016WaveNetAG}, which inspires researchers to explore its effectiveness in action segmentation~\cite{ColinLea2016TemporalCN}. Temporal Deformable Residual Networks (TDRN)~\cite{PengLei2018TemporalDR} further introduces deformable convolutions and residual streams to realize fine-grained recognition. Recently, multi-stage TCN~\cite{farha2019ms} introduces a multi-stage cascaded structure, which inspires us to introduce a cascade module to the OAD task. Last but not least, researchers~\cite{FangqiuYi2022ASFormerTF, du2022efficient, wang2022cross} pay more attention on the Transformer for action segmentation.
The most significant difference between online action detection and offline action segmentation is whether future information is accessible, which leads to differences in their inference mechanisms. Our work attempts to establish the relationship between these two tasks, to obtain performance improvements on the online action detection task.

\subsection{Offline Action Detection} 
% 这一工作存在的难点是什么? 这一任务的难点是什么? 基于这一难点(当然可能没有难点, 把近期的工作进行一个稍微的整理即可)，近期衍生出了哪些解决方案? 应该如何解决?  
% Colar: Effective and Efficient Online Action Detection by Consulting Exemplars
% ActionFormer: 从One Stage和Two Stage的角度来深度介绍; 
Similar to offline action segmentation, offline action detection aims to detect the action boundaries and identify their categories after fully observing untrimmed videos. A line of previous work can be categorized as one-stage methods~\cite{WeiLiu2016SSDSS, TianweiLin2017SingleST, RunhaoZeng2020DenseRN,ChumingLin2021LearningSB,zhang2022actionformer} and two-stage methods~\cite{ShyamalBuch2017SSTST,TianweiLin2019BMNBN,RunhaoZeng2019GraphCN,qing2021temporal,zhu2021enriching}. 
The two-stage approaches generate multiple candidate action proposals and perform categorical identification and boundary regression.
Previous two-stage work~\cite{VictorEscorcia2016DAPsDA,zhao2020bottom,YueranBai2020BoundaryCG,JingTan2021RelaxedTD,YuanLiu2018MultigranularityGF,PeisenZhao2020BottomUpTA} most focuses on generating action proposals. For examples,~\cite{VictorEscorcia2016DAPsDA} generates the proposals based on anchor classification and ~\cite{YuanLiu2018MultigranularityGF,PeisenZhao2020BottomUpTA} generates based on the frame-level probabilistic sequence. Besides, the recent Transformer-based~\cite{qing2021temporal,zhu2021enriching} and GCN-based methods~\cite{ChenZhao2020VideoSG,MengmengXu2019GTADSL} incorporate temporal context by exploiting the relations among the proposals. 
However, the one-stage approach~\cite{zhang2022actionformer,TianweiLin2017SingleST,ShyamalBuch2017EndtoEndST,LongFuchen2019GaussianTA,ChumingLin2021LearningSB} localize actions without proposals, roughly divided into anchor-based approach~\cite{ShyamalBuch2017EndtoEndST,LongFuchen2019GaussianTA} and anchor-free approach~\cite{ChumingLin2021LearningSB,zhang2022actionformer}. 
%\textcolor{red}{~\cite{WeiLiu2016SSDSS} draws on the idea of one-stage object detector~\cite{WeiLiu2016SSDSS} and proposes a one-stage anchor-based method. The anchor-free approach~\cite{zhang2022actionformer} generally localize the action segments based on the frame-level predictions like sequence labeling.} 
Lin~\textit{et al}. design the first one-stage anchor-based method SSAD~\cite{TianweiLin2017SingleST}, motivated by the one-stage object detection approaches~\cite{WeiLiu2016SSDSS}. Zhang~\textit{et al}. propose an one-stage anchor-free method by frame wise classification using Transformer.

% 补充一下LSTR的结果
\subsection{Transformer}
Transformer \cite{vaswani2017attention} is first proposed to handle the tasks in natural language processing, and it has made breakthrough progress. This success has inspired many researchers to explore the application of Transformers in the visual domain \cite{jaegle2021perceiver,neimark2021video,sharir2021image,li2021vidtr,bertasius2021space,arnab2021vivit,nawhal2021activity,tan2021relaxed,dosovitskiy2020image,carion2020end}. ViT \cite{dosovitskiy2020image} is the first Transformer-based architecture for image classification and it achieves promising performance. For object detection, DETR  \cite{carion2020end} abandons traditional hand-designed modules and replaces them with Transformers. There are also some attempts \cite{neimark2021video,sharir2021image,li2021vidtr,bertasius2021space,arnab2021vivit,nawhal2021activity,tan2021relaxed} in modeling temporal dependencies of video using Transformer.

Though Transformer has achieved remarkable performance on many vision tasks, the extraordinary demands of computing resources limit its further application. 
% To solve this problem, recent work attempts to introduce sparse attention~\cite{liu2021swin,wang2021kvt} or pruning strategy~\cite{chen2021chasing} into the transformer.
% There are other works~\cite{rao2021dynamicvit,tang2021patch} focus on dropping the unimportant tokens that have little impact on the performance to reduce unnecessary computation.
% Besides, Perceiver~\cite{jaegle2021perceiver} decouples the network depth from the sequence length by alleviating the computational complexity that grows quadratically with the sequence length, which enables to build very deep models. HiP~\cite{carreira2022hierarchical} introduces locality in Perceiver, which saves computation and expands the model's ability to handle high-resolution inputs.
Recent work has made many efforts on the following aspects: i) sliding windows~\cite{parmar2018image, child2019generating, ho2019axial, liu2021swin}. For example, Swin Transformer~\cite{liu2021swin} advocates local window attention to reduce the quadratic complexity dramatically.
ii) low rank~\cite{wang2020linformer, xiong2021nystromformer, tay2021synthesizer}. Linformer~\cite{wang2020linformer} projects keys and values into two embedding with an ultra-low dimension, so computation costs will be significantly reduced when queries attend to these embedding.    
iii) memory~\cite{rae2019compressive, lee2019set}. For instance, Set Transformer~\cite{lee2019set} adopts extra memories to remember the attended features for distant tokens. 
iv) kernel-based approximation~\cite{choromanski2020rethinking, katharopoulos2020transformers, peng2021random}. These works have explored various efficient mechanisms to alleviate the undesirable quadratic complexity in Transformers. Considering the FIFO pipeline in streaming videos, a straightforward idea to avoid redundant computation is to tackle the head and tail while reusing the intermediate already computed, which is the prerequisite to support complex modules such as hierarchical stages or even cascade in this paper.

\section{Our Approach}
\subsection{Overview}
% 介绍任务
OAD aims to identify the action based on long historical information without accessing future information. Therefore, improving the ability to model historical details as much as possible is critical while ensuring a limited amount of computation. We advocate a novel window-based mechanism, especially for streaming video understanding with this principle. %是否需要展开说时如何加速的训练的? 
During online streaming inference, we merely model the changes in history by updating the latest and the oldest historical representations in one window and remaining the intermediate ones. This operation allows us to perform hierarchical modeling and cascade refinement within only one window, which means we have a low computation cost for streaming inference when partially updating the compressed history. 

\noindent\textbf{Task Definition.} Mathematically, given a streaming video, the proposed method aims to yield an action probability vector $\hat{\boldsymbol{y}_t} \in \mathbb{R}^{N_a}$ of the current frame $\boldsymbol{x}_t\in \mathbb{R}^d$ at time $t$ based on $\{\boldsymbol{x}_1, \dots, \boldsymbol{x}_t\}$, where $d$ is the number of channels. Here $N_a$ is the total number of actions. Following the existing setting, all the frames $\{\boldsymbol{x}_1, \boldsymbol{x}_2, ...\}$ are represented by feature vectors extracted from a pre-trained network.

\begin{figure*}
\centering
%\hspace{10mm}
\includegraphics[height=9cm]{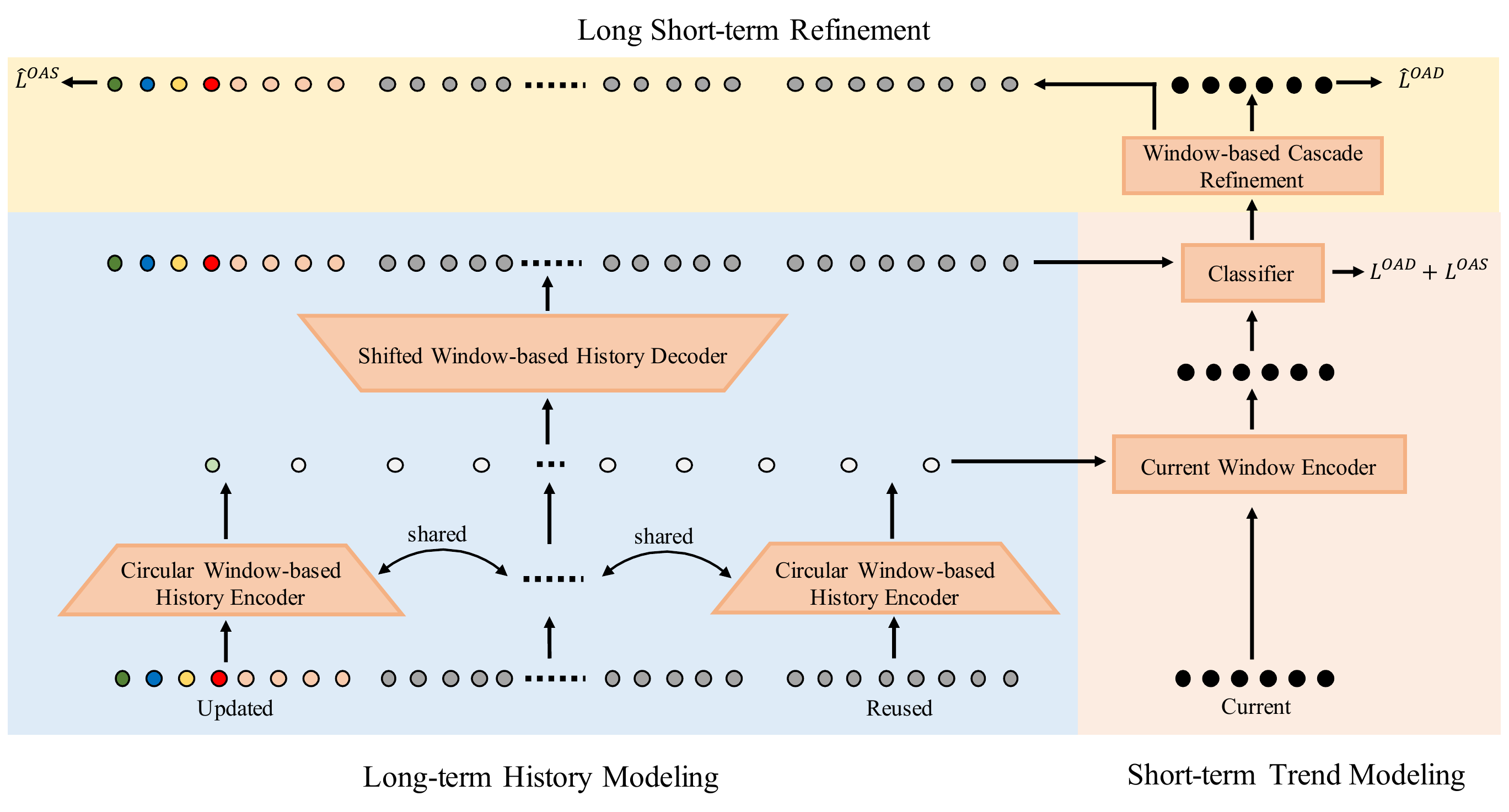}
\caption{Overview of the proposed CWC-Trans for online action detection. The CWC-Trans consists of three key components: long-term history modeling, short-term trend modeling, and long short-term refinement. For the details of each component, please refer to the following subsections in the method.} 
\label{fig:overview}
\end{figure*}

In this paper, we propose a novel Transformer called Circular Window-based Cascade Transformer (\textbf{CWC-Trans}). The architecture of our model is illustrated in Figure~\ref{fig:overview}. To better understand the whole framework, we briefly introduce its three key components: long-term history modeling, short-term trend modeling, and long short-term refinement. Then we demonstrate the training strategy which uses a multi-task training mechanism and two kinds of online inference.

\subsection{Long-term History Modeling}
 In order to handle the varied history, we follow the previous work~\cite{xu2021long} to limit the history to $\{\boldsymbol{x}_{t - m_L - m_S + 1}, \dots, \boldsymbol{x}_t\}$, where $m_L$ and $m_S$ denote the number of frames in the long-term history and short-term trend, respectively. The long-term historical feature bank embodies long-range casual across time. Thus, it is supposed to be compressed to a limited number of representatives so that it can be abstracted and used highly efficiently. Specifically, we divide it into $N_w$ windows $E=\{\boldsymbol{x}_{t - m_S - nw + 1}, \dots, \boldsymbol{x}_{t - m_S - (n-1)w}\}_{n=1}^{N_w}$, where $w = \frac{m_L}{N_w}$ represents the window size. Then each window is fed into a hierarchical attention module, and the final output of each window is represented by a vector $\boldsymbol{c}_{n}$. So the compressed historical information is finalized as $\boldsymbol{C} = \{\boldsymbol{c}_1, \boldsymbol{c}_2, \dots, \boldsymbol{c}_{N_w}\}$. Note that we embed the latest and the oldest historical representations in the same window, which guarantees an efficient updating mechanism for online inference. During training, we append a decoder on $\boldsymbol{C}$ and introduce an auxiliary task, Offline Action Segmentation (OAS), to learn compact and informative encoding $\boldsymbol{C}$. Note that the decoder used for training does not introduce extra computation costs during testing, no matter how complex it is. We denote the outputs of the auxiliary task as $\hat{\boldsymbol{Y}}_{L}^{\prime} = \{\hat{\boldsymbol{y}}_{t-m_L-m_S+1}^{\prime}, \dots, \hat{\boldsymbol{y}}_{t-m_S}^{\prime}\}$. The detailed structure of Long-term History Modeling is illustrated in Figure~\ref{fig:mtsm}. We introduce each component in detail as follows.

\subsubsection{Circular Window-based History Encoder (CWHE)}

\textbf{History Feature Dimension Reduction.} % How to analysis? 
% 为什么这样做? 这样做有什么好处? (如何理解?) 
% 降低了多少计算量、内存消耗
% 写一下Attention的复杂量计算公式? 
% Current vs History 
In videos, one obvious property is that adjacent frames share redundant spatial information, especially in static cameras. Therefore, we lower the channel capacity of the long historical representations by linear projection. More importantly, this dimension reduction can significantly reduce the computation for online inference thanks to the large $m_L$. %This is what makes the Fast pathway more computation-effective than the Slow pathway. 
On the other hand, the short-term trend is much more likely associated with the current frame as they share the same action most often. It means remaining channel capacity during projection for trend features will preserve recent contexts as much as possible. Besides, the limited number of trend frames does not yield a high computation cost. Such an operation also obeys the setting in Slowfast~\cite{feichtenhofer2019slowfast} where the fast pathway uses a lower channel capacity than the slow one.
We formulate this operation as follows: 
\begin{equation}
\begin{aligned}
\label{eq:feature_projection}
\boldsymbol{e}_{i} &={\boldsymbol{W}_L} \boldsymbol{x}_i, && t_{t-m_S-m_L+1} \leq i \leq t_{t-m_S}  \\
\boldsymbol{e}_{i} &={\boldsymbol{W}_S} \boldsymbol{x}_i, && t_{t-m_S+1} \leq i \leq t
\end{aligned}
\end{equation}
where $\boldsymbol{W}_{L} \in \mathbb{R}^{d_{L} \times d}, \boldsymbol{W}_{S} \in \mathbb{R}^{d_{S} \times d}, $ and $d_{L}<d_{S}$. 

\noindent\textbf{Non-overlapped Window Partition and Partial Updating.}
%% 为什么要划分窗口(计算复杂度的指数关系?), 划分窗口有什么好处; Spacially 如何划分窗口的, 给出一些数学公式, 窗口划分公式、Attention计算公式; position embedding 的设计? Window Partation
%% 灵感是怎么来的(Swin \cite{liu2021swin} 在各项视觉任务上都取得了很大的成功; 这启发我们挖掘序列中的局部依赖关系。)? -> 然而, 基于Swin-Transformer往往会带来巨大的计算量，没有办法直接应用于在线动作检测任务重。通过分析我们发现，如果通过一种合理的方式划分窗口(更新机制), 是能够进一步减轻计算量的; 
% The Window Based Attention are Local operation, which may lack of global , So we apply the self-Attention on the compressed historical sequences 以建立序列元素的长期依赖. 
Recall that we split the historical feature bank into multiple non-overlapped windows. Here we explain why we need a non-overlapped partition rather than overlapped one. To begin with, the most important reason is the efficient partial updating during online inference. It has to update all the relevant windows or even all when going to a deep network if we adopt an overlapped window. This situation also happens during hierarchical attention in history if we allow window correlation such as window shift or halo to increase the receptive field.
On the other hand, global attention can be achieved when the short-term trend attends to the compressed history. Thus, we isolate all the non-overlapped windows during the hierarchical history abstract. Besides, the non-overlapping operation enables us to reuse most of the windows already computed in the last step and partially update only one circular window at the moment. We only need to drop the oldest history frame and move the latest one to the circular window for every moment in streaming videos. The updating function $g$ of windows can be expressed by the following equations:
\begin{equation}
g(\boldsymbol{E}_n) = \\
\begin{cases} 
\boldsymbol{\vec{E}}_{n} & \text{ if } n = N_{w} \\ 
\boldsymbol{E}_{n} & \text{ if } n \neq N_{w}
\end{cases} \\
\end{equation}
where $\boldsymbol{E}_{n}=\{\boldsymbol{e}_{t-m_S - nw+1}, \dots, \boldsymbol{e}_{t-m_S-(n-1)w}\}$, $\boldsymbol{\vec{E}}_{n}=\{\boldsymbol{e}_{t-m_S+1}, \boldsymbol{e}_{t-m_S-nw+2},\dots, \boldsymbol{e}_{t-m_S-(n-1)w}\}$. 

\begin{figure*}[t]
\centering 
\includegraphics[width=18cm]{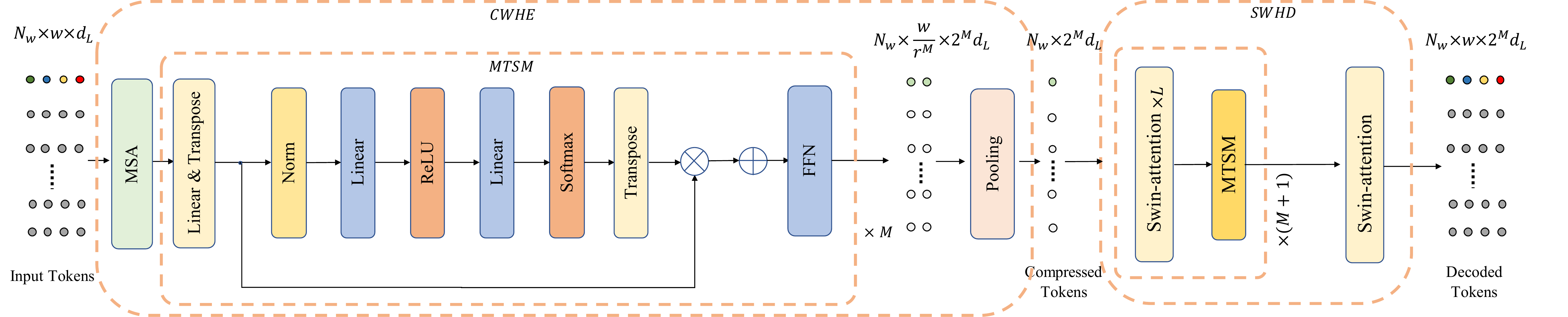}
%\caption{\textcolor{blue}{Illustration of Circular Window-based History Encoder (CWHE), including a Multi-head Token Slim Module (MTSM) and Shift Window-based History Decoder (SWHD). $\otimes $ represents matrix multiplication between the token representations and the attention matrix, and $\oplus$ represents the concatenation of all the head outputs.}}
\caption{Illustration of Long-term History Modeling. It consists of Circular Window-based History Encoder (CWHE) and Shift Window-based History Decoder (SWHD). $\otimes $ represents matrix multiplication between the token representations and the attention matrix, and $\oplus$ represents the concatenation of all the head outputs.}
\label{fig:mtsm}
\end{figure*}

\noindent\textbf{Hierarchical Window Attention with Multi-head Token Slim Modules.}
Afterwards, we employ weight-shared self-attention in each historical window to capture their local temporal dependencies. 
%% 是否需要提一句，为什么不需要引入Position Embedding? \先不说, 最后在修正
% 介绍一下MSA的优势和优点, 然后在紧接着列出求解公式.
The Multi-head Self-Attention (MSA) mechanism is a fundamental component of Transformer, effectively capturing the dependencies between sequence elements. To be specific, self-attention enables interactions between every two tokens, which can build rich contextual semantic information based on the tokens' characteristics. And the multi-head incorporates multiple patterns among tokens, enhancing their expressiveness. The formulation of $i^{th}$ head self-attention on $n^{th}$ history window $W_n$ is shown below: 

\begin{equation}
\begin{aligned}
% V_{w_n}^{\prime} &=\operatorname{LayerNorm}\left(V_{w_n}\right) \\
\boldsymbol{Q}^{i}_n &=g(\boldsymbol{E}_n) \boldsymbol{W}^{i}_{q} \\ 
\boldsymbol{K}^{i}_n &=g(\boldsymbol{E}_n) \boldsymbol{W}^{i}_{k} \\ 
\boldsymbol{V}^{i}_n &=g(\boldsymbol{E}_n) \boldsymbol{W}^{i}_{v} \\
\boldsymbol{E}^{\prime i}_n &=\operatorname{Softmax}\left(\frac{\boldsymbol{Q}^{i}_n {\boldsymbol{K}_{i}^n}^{\top}}{\sqrt{d_{k}}}\right) \boldsymbol{V}^{i}_n \\
\end{aligned}    
\end{equation}
where $\boldsymbol{W}^{i}_{q}, \boldsymbol{W}^{i}_{k}, \boldsymbol{W}^{i}_{v}$ are the linear projection matrix to yield the query $\boldsymbol{Q}^{i}_n$, key $\boldsymbol{K}^{i}_n$, and value $\boldsymbol{V}^{i}_n$. ${d_k}$ is the scaling factor that can accelerate convergence. $\boldsymbol{E}^{\prime i}_n$ represents the output of the $i^{th}$ head.
After that, we concatenate the output of $N_{h}$ heads and feed it into a fully-connected layer $f$ for the final output $\boldsymbol{E}_n^{\prime}$, as shown below.

\begin{equation}
\boldsymbol{E}^{\prime}_n={f}\left([\boldsymbol{E}^{\prime 1}_n, \boldsymbol{E}^{\prime 2}_n, \ldots, \boldsymbol{E}^{\prime N_h}_n]\right) \\
\end{equation}

% 引入TSM之后应该说，传统的TSM, ....., 不适合我们这种渐进式的压缩。为此，我们提出了TSM.
% 他是如何Work的~公式符号,
% 别人了解你的工作之后, 就可以总结一下，你的方法的好处。
% 原始的TSM缺点（没有改变序列的Channel;? 整体容量有所下降, (给个图?)
%% 提出了函数降维的必要性; 同时, 提出提升特征容量的优点, 基于已有的方法进行改进? 
%% 更有效的降维方式? [与Patch Mergeing相对应;] 
% 因为Decoder的计算量与历史序列的长度高度相关, 因此需要一些有效的序列压缩策略来降低历史序列的长度, 具体地～～;
% Since the computation of the decoder is highly correlated with the length of the history sequence, 
Since the history will be queried by the current window, some efficient sequence compression strategies are needed to reduce the length at each hierarchical stage. For example, LSTR~\cite{xu2021long} adopts cross-attention to aggregate the long historical sequences to a few tokens using a data-agnostic mechanism like NetVLAD~\cite{arandjelovic2016netvlad}. However, it does not preserve the sequence order once applied, and a significant compression rate will cause dramatic information loss. Besides, Patch Merging~\cite{liu2021swin} is another way to compress the number of tokens, accumulating the local patch by stacking them in the channel. However, the flattening operation lacks dynamic token selection, where the most critical components are expected to be preserved. A recent work~\cite{zong2021self} designs a self-attention-like module named Token Slimming Module (TSM) that enables dynamic token aggregation. Although the TSM solves the two problems mentioned above, the single slimming matrix in TSM does not sufficiently explore the representation capacity in the tokens. Inspired by the MSA, we propose a novel Multi-head Token Slimming Module (MTSM), as shown in Figure~\ref{fig:mtsm}. Specifically, the compressed token of the $i^{th}$ head in $N_{h\prime}$ heads can be formulated as follows. 
% \begin{equation}
% \begin{aligned}
% &A_i =\operatorname{Softmax}\left({W_{qi} \sigma\left(\mathbf{V_{w_n}} W_{k}\right)^{T}}\right) \\
% &V_{iw_n}^{\prime} ={{A_i} {V_{iw_n}}}
% \end{aligned}
% \end{equation}

% %看这个例子,然后history decoder基于C_n继续写下去
% \begin{equation}
% \begin{aligned}
% &\boldsymbol{A}_n =\operatorname{Softmax}\left({\boldsymbol{W}_{r} \sigma\left(\boldsymbol{H_n} \boldsymbol{W}_{c}\right)^{T}}\right) \\
% &\boldsymbol{C}_{n} ={\boldsymbol{A}_i \boldsymbol{H_n}}
% \end{aligned}
% \end{equation}
\begin{equation}
\begin{aligned}
&\boldsymbol{T}^{i}_{n} ={\boldsymbol{E_n^{\prime}} \boldsymbol{W}^i_{h}} \\
&\boldsymbol{E}^{\prime\prime i}_{n} =\operatorname{Softmax}\left({\boldsymbol{W}_{r} \sigma\left(\boldsymbol{T}^{i}_{n} \boldsymbol{W}_{c}\right)^{T}}\right) \boldsymbol{T}^{i}_{n} \\
% &\boldsymbol{H}^{\prime i}_{n} ={\boldsymbol{A}^{i}_n \boldsymbol{V}^{\prime i}_{n}} \\
%&\boldsymbol{C}_n={f}\left([\boldsymbol{H}^{\prime 1}_n, \boldsymbol{H}^{\prime 2}_n, \ldots, \boldsymbol{H}^{\prime N_h}_n]\right) \\
\end{aligned} 
\end{equation}
where $\boldsymbol{W}_{h}^i$ is used for head partition in the channel, and it yields the feature $\boldsymbol{T}_{n}^i$ of the $i^{th}$ head. $\boldsymbol{W}_{c}$ is a bottleneck matrix to reduce the channel, while $\boldsymbol{W}_{r}$ is used for token reduction in each window. $\sigma$ represents the activation function ReLU. $\boldsymbol{E}^{\prime\prime i}_{n}$ is the output of the $i^{th}$ head.

% \begin{figure}[t]
% \centering
% \includegraphics[width=8.0cm]{figures/ghd.pdf}
% \caption{Illustration of Shift Window-based History Decoder (SWHD). The Swin-attention module is used to enhance the interactions among tokens, and the MTSM up-samples the compressed input tokens. Through multi-stage decoding, the historical sequence resolution will be fully recovered.}
% \label{fig:ghd}
% \end{figure}

Afterwards, we concatenate the outputs of all the heads and feed them into a fully connected layer $f^{\prime}$.
\begin{equation}
\boldsymbol{E}_n^{\prime\prime}={f^{\prime}}\left([\boldsymbol{E}^{\prime\prime 1}_n, \boldsymbol{E}^{\prime\prime 2}_n, \ldots, \boldsymbol{E}^{\prime\prime N_{h^{\prime}}}_n]\right)
\end{equation}
where $\boldsymbol{E}_n^{\prime \prime} \in {R}^{\frac{w}{r} \times {2d}}$ represents the compressed tokens with greater channel capacity. Here, $r$ is the token reduction rate. We repeat window attention with MTSM by $M$ times to form the hierarchical window attention. It dramatically reduces the number of tokens. Finally, we use mean pooling to squeeze the window into one vector, namely $\boldsymbol{c}_n$. Note that we deploy one attention layer for efficiency during online inference. We also find that positional encoding is unnecessary for the history encoder inputs. We believe that history provides a global view for the current frame classification, so temporal positions may not be useful as they are far away from now. Last but not least, we employ a few global self-attention layers on $\boldsymbol{C}$ to finalize the global representation that is absent in the previous window attention. 

\begin{figure}[t]
\centering
\includegraphics[height=4.5cm]{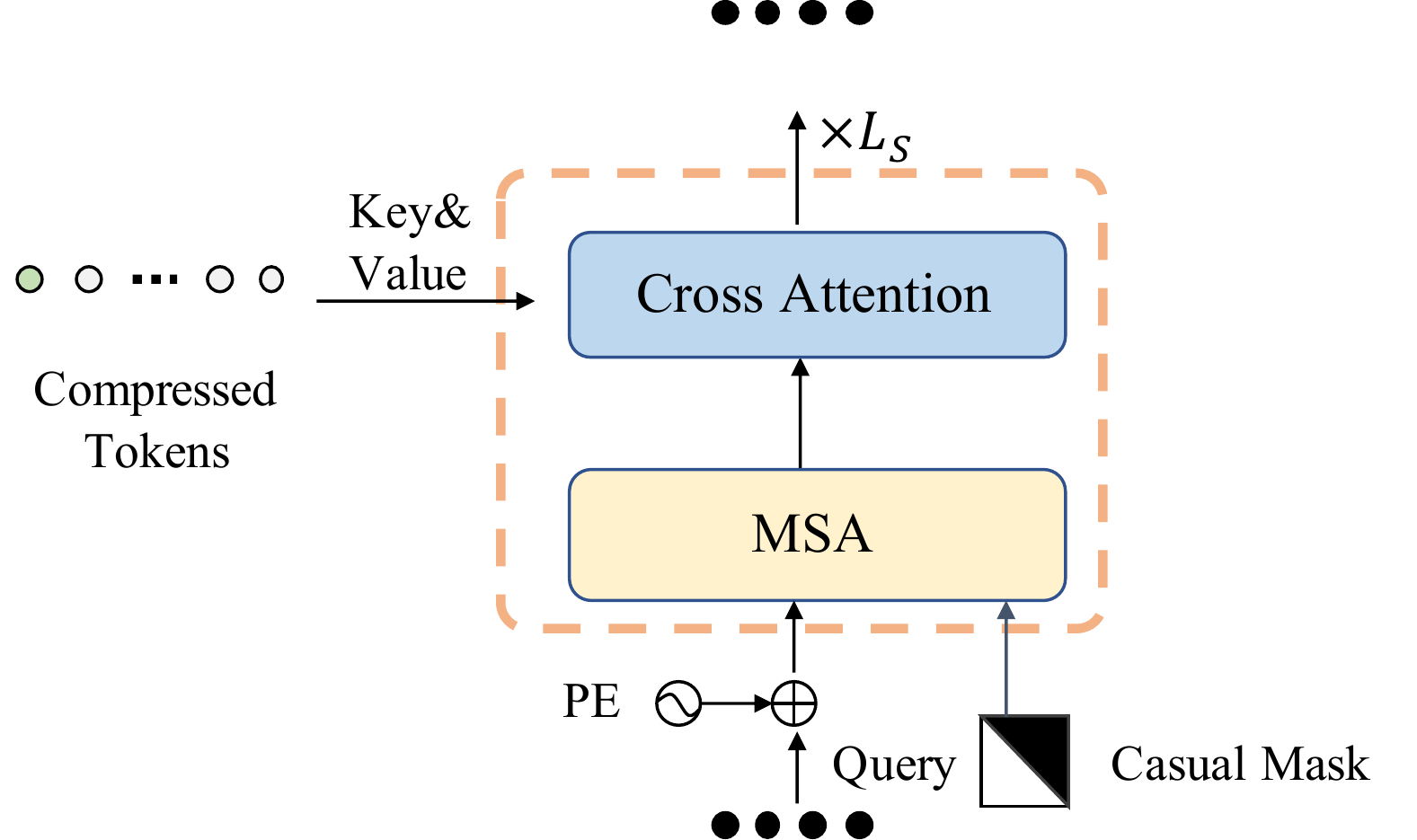}
\caption{Illustration of Current Window Encoder (CWE). The trend tokens are first fed into one self-attention layer and then are incorporated with the long historical information by cross-attention.}
\label{fig:cwe}
\end{figure}

\subsubsection{Shifted Window-based History Decoder (SWHD)} 
\label{sec:history_decoder}
The historical sequence compression will inevitably result in the loss of some critical action cues. However, the existing work LSTR~\cite{xu2021long} and GateHUB~\cite{chen2022gatehub} do not explicitly preserve all the historical information as much as possible, leading to a suboptimal compression due to the sole OAD supervision. Therefore, we introduce OAS as an auxiliary training task to alleviate this problem. The training details will be introduced in Section~\ref{sec:mtl} and here we demonstrate the decoder architecture on the compression $\boldsymbol{C}$. We design a window-based history decoder, as shown in Figure~\ref{fig:mtsm}. Unlike the encoder used during online inference, the auxiliary task is only available during training. Thus, window shift is allowed to enlarge the receptive field for better action segmentation. Specifically, we apply one decoder module with $L$ Swin~\cite{liu2021swin} layers and MTSM at each stage. In contrast to the MTSM in the encoder, MTSM used in the decoder intends to up-sample the tokens in windows, which means the token expansion rate $r^{\prime}$ is reversely proportional to $r$. After that, an extra Swin layer is applied to align back to the space before. To recover the full temporal resolution, we need to use the decoder module by $M+1$ times as we apply hierarchical window attention by $M$ times plus one mean pooling in the history encoder. We plug a classifier $\phi$ at the end to realize OAS, which produces the coarse probabilities $\hat{\boldsymbol{Y}}_{L}^{\prime}$.

\subsection{Short-term Trend Modeling} 
\label{sec:current_decoder}
On the other hand, the short-term trend as a strong motion cue contains fewer frames ($m_S \ll m_L$) than history. Thus, we retain its limited resolution. The trend first performs causal attention where each one only attends to itself and the previous. The attended trend then attends to the compressed clusters $\boldsymbol{C}$ to fuse all the information and predict the probabilities $\hat{\boldsymbol{Y}}_{S}^{\prime} = \{\hat{\boldsymbol{y}}_{t-m_S+1}^{\prime}, \dots, \hat{\boldsymbol{y}}_{t}^{\prime}\}$. 

% 因为短期历史能够对识别当前帧提供更多有效的信息，因此, 我们从所有的历史序列中分出一部分来, 采用一种无压缩的方式(保证时空分辨率)，直接与当前帧建立交互关系，并且利用压缩后的历史信息得出最终的概率预测。
% as shown in figure~xxx具体地! self-attention with casual Mask, model the most related context. Then use cross-attention to retrieve the information from the compressed historical information. It outputs $m_{S}$ probability vectors $\left\{\mathbf{p}_{T}, \cdots, \mathbf{p}_{T-m_{S}+1}\right\} \in[0,1]^{K+1}$, each $\mathbf{p}_{t}$ representing the predicted probability distribution of $K$ action categories and one "background" class at time $t$. 
%在测试的时候, 我们只取当前帧的预测(因为能够更充分地利用当前帧地信息);  
\subsubsection{Current Window Encoder (CWE)}
The observations $\left\{\boldsymbol{x}_{t-m_{S}+1}, \ldots, \boldsymbol{x}_{t}\right\}$ as a short-term trend is significantly informative to the current frame $x_t$, which brings the most relevant context. Therefore, we isolate this part from history and treat it as a current window. The tokens in this window interact with each other using an uncompressed approach which maintains its original temporal resolution and high channel capacity. We demonstrate the current window encoder in Figure~\ref{fig:cwe}. After channel reduction using Equation~\ref{eq:feature_projection}, we add sinusoidal position encoding $\boldsymbol{E}_{pos}$ to $\left\{\boldsymbol{e}_{t-m_{S}+1}, \ldots, \boldsymbol{e}_{t}\right\}$ and then feed them into $L_S$ MSA layers with a casual mask $\boldsymbol{\mathcal{M}}$ to model the temporal dependencies for each frame. The causal mask ensures that each frame will only attend to the frames from the past to itself. After that, we take the current representations self-attended to cross attend to the compressed historical tokens $\boldsymbol{C}$. However, the window partition on the ordered history will not simulate the circular window. Thus, we shift the history bank multiple times to better align the current observations and their corresponding history. It results in different copies of the history bank, most of which overlap with the current window. The cross attention finally yields the predictions $\{\hat{\boldsymbol{y}}_{t-m_S+1}^{\prime}, \dots, \hat{\boldsymbol{y}}_{t}^{\prime}\}$ of the current window by appending the classifier $\phi$ shared with action segmentation. We align these two tasks as they share the common target, namely per-frame action classification, except for the accessibility to the future. Such a scheme functions as a data augmentation for online action detection. During online inference, only the prediction $\hat{y}_{t}^{\prime}$ of the latest frame is used for online action detection. 
\begin{figure}[t]
\centering
\includegraphics[height=4.2cm]{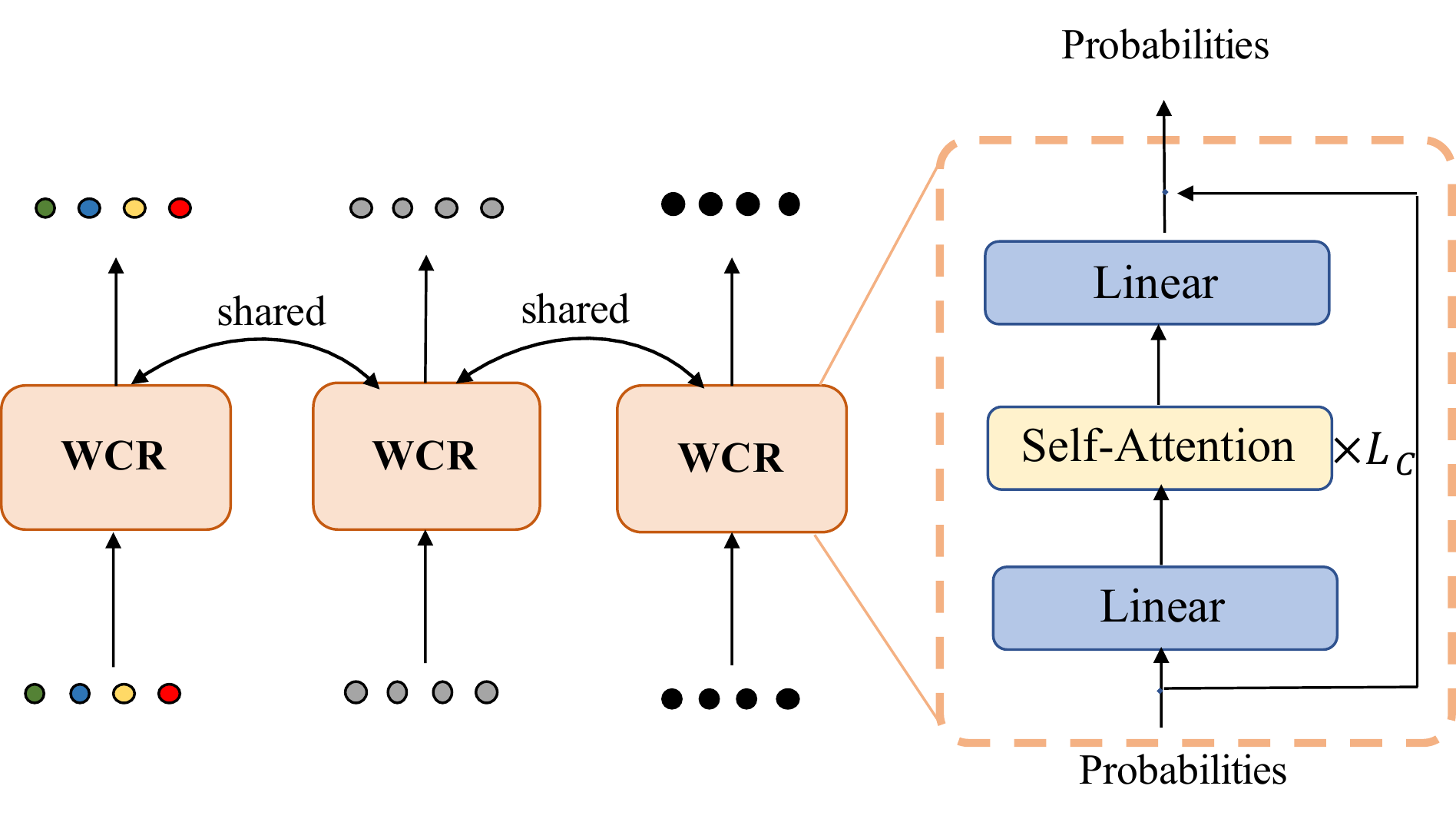}
\centering
\caption{Illustration of Window-based Cascade Refinement (WCR) module. The left means the cascade module is shared by the history and trend tokens. The right is the details of the cascade module.}
\label{fig:cwhe:crm}
\end{figure}

\subsubsection{Action Classifier}
Here we highlight the action classifier realized by a linear layers for two reasons. First, its outputs are the inputs of the long short-term refinement module, which the effectiveness of refinement on the coarse probabilities has been well verified by OAS~\cite{farha2019ms}. Second, the action classifier is shared by OAD and OAS, as shown in Figure~\ref{fig:overview}.

\subsection{Long Short-term Refinement}
\textbf{Window-based Cascade Refinement (WCR).} Note that OAD on the short-term trend and OAS on the long-term history share the same target, \ie, assigning an action label to each frame, except for the accessibility to the future. As actions own continuity near boundaries which leads to ambiguity, a classical solution in OAS to alleviate this problem is label refinement. Thus, we further feed the coarse probabilities $\hat{\boldsymbol{Y}}_{L + S}^{\prime} = \hat{\boldsymbol{Y}}_{L}^{\prime} \bigcup \hat{\boldsymbol{Y}}_{S}^{\prime}$ of both long-term and short-term information into a shared cascade module that uses several Transformer encoder layers to yield final predictions $\hat{\boldsymbol{Y}}_{L+S} = \{\hat{\boldsymbol{y}}_{t-m_L-m_S+1}, \dots, \hat{\boldsymbol{y}}_{t-m_S}, \hat{\boldsymbol{y}}_{t-m_S+1}, \dots, \hat{\boldsymbol{y}}_{t}\}$. Again, the refinement of long-term history will not bring extra computation costs during testing as it functions like a regularization. 

The mapping from features to probabilities always yields discontinuous predictions cross time, as indicated in offline action segmentation~\cite{farha2019ms}. We also observe such a drawback in online action detection. One effective remedy is to append a cascade refinement upon the predicted probabilities. For example, MS-TCN~\cite{farha2019ms} proposes a multi-stage temporal convolution network, where each stage takes probabilities from the previous stage as inputs for the next stage and smooths the coarse predictions gradually. However, we also need to be concerned about the computation complexity due to the real-time demand. To this end, we design a lightweight cascade refinement module consisting of $L_C$ self-attention layers to refine the current predictions $\hat{\boldsymbol{Y}}_{S}^{\prime}$, as shown in Figure~\ref{fig:cwhe:crm}. To accelerate the convergence, we add a shortcut between inputs and outputs at each cascade refinement as both are probabilities. Similar to the shared classifier $\phi$, we reuse this cascade refinement on OAS outputs $\hat{\boldsymbol{Y}}_{L}^{\prime}$ where refinement is performed in each single history window. The outputs of OAS act as data augmentation to train a holistic refinement network for $\hat{\boldsymbol{Y}}_{S}^{\prime}$.
Meanwhile, the cascade used on OAS will not bring extra computation costs during online inference. To better align OAD and OAS in the feature space of the cascade, we apply the causal mask on the current window but also on every history window. That means each coarse prediction in a history window will attend to the previous and itself. Overall, the window-based cascade refinement $\mathcal{F}$ can be formulated as follows. 

\begin{equation}
\begin{aligned}
&\boldsymbol{P}^{0}=\hat{\boldsymbol{Y}}_{L + S}^{\prime} \\
&\boldsymbol{P}^{i}=\mathcal{F}\left(\boldsymbol{P}^{i-1}, \boldsymbol{\mathcal{M}}\right)
\end{aligned}
\end{equation}
where $\boldsymbol{\mathcal{M}}$ represents the casual mask, which is the same as we discussed in Section~\ref{sec:current_decoder}. We can repeat the cascade refinement multiple times, but it will bring a significant computation burden. We empirically find that even one cascade refinement is enough to achieve promising performance. Thus, the proposed refinement is lightweight and effective.

\subsection{Training and Online Inference}
\label{sec:mtl}
% 写清楚Motivation, 为什么要做如此的设计? 
%之前的工做把,OAD和OAS当作是两个不同的任务（相同的话从不同的角度来写，这样总是有干货：角度1 XX; 角度2: ），（然而，OAS是一种十分有效的视频理解方法, Dense Predictions of each frame) 动作分割任务, 和压缩关系十分密切; 两者一起学习的话，动作分割的损失
% 只专注于单个模型可能会忽略一些相关任务中可能提升目标任务的潜在信息，通过进行一定程度的共享不同任务之间的参数，可能会使原任务泛化更好。
% 什么是多任务学习?
%Multi-Task Learning(MTL) aims to achieve consensus between multiple relevant tasks, while it may improve the generalizability of a single task. In MTL, the auxiliary task would provide
Although the inference mechanisms of OAS and OAD are different, their common targets can still enhance each other, which has not been well studied in previous literature. Motivated by this observation, we introduce OAS as an auxiliary task to OAD for compact and informative compression of long-term historical information. We present the objective functions for end-to-end training CWC-Trans in the following subsections.

%% 由于历史序列非常长, 可以当作离线视频（本质上都是对压缩的Token建立）, 因此可以引入基于AS的多任务学习，辅助压缩过程产生更discriminative representations.
%% 简单提一下这个因素, （Local Attention比较大），动作割裂等都是其中的具体表现, 以此为出发点介绍引入Action Segmentation的目的(简单说一句); 
%% Besides, the proposed window-based attention may introduced wrong(引入错误的先验[动作的割裂、]), To alleviate this biase, we Introduce Action Segmentation Loss.  
%% 可以用作降维的参考?)简言之就是借用AS中的一些通用Global设计(CA?)?~Tricks(如何设计对矩阵进行变换;) 
\subsubsection{Training Objective Functions}
\textbf{Current Classification Loss}.
OAD aims to classify the current frame correctly. However, regularization on other frames in the current window plays a useful supervision signal as indicated by LSTR~\cite{xu2021long}. So we employ a softmax with cross-entropy loss over all the frames in the current window, demonstrated as follows.
\begin{equation}
\mathcal{L}^{\mathrm{OAD}}=-\sum_{i=t-m_S+1}^{t} \sum_{j=1}^{N_a} \boldsymbol{y}_{i, j} \log \hat{\boldsymbol{y}}_{i, j}^{\prime}
\end{equation}
where $\boldsymbol{y}_i$ is the ground-truth of the $i^{th}$ frame, which is one-hot encoding.

% 看到未来的先验知识; 
%% this often results in a solution that generalizes better.
% 寻找一个有效的压缩方式是很难得，通过单纯的Online监督方式, 往往会导致（单向的监督，很难建立起真正有效的监督方式, 通过Action Segmentation这种直接监督的方式; may产生更有效的压缩Token;）
% Action Segmentation % 可以用来捕捉需贴动作之间高阶依赖关系, 通常很难直接建立by sample supervision of the main task. 
% 比如只存在很高阶的相关性，或被其他因素抑制
% 理论上来说，通过引入动作分割的损失能够极大的提升我们的模型的性能; 
% 可以推导出表达能力更巧的特征; 
\noindent\textbf{Action Segmentation Loss}.
As demonstrated in Section~\ref{sec:history_decoder}, the history decoder projects the compressed vectors back to the vanilla resolution. Analogous to the current classification loss, the action segmentation loss is defined as:

\begin{equation}
\mathcal{L}^{\mathrm{OAS}}=-\sum_{i=t-m_S-m_L+1}^{t-m_S} \sum_{j=1}^{N_a} \boldsymbol{y}_{i, j} \log \hat{\boldsymbol{y}}_{i, j}^{\prime}
\end{equation}

\noindent\textbf{Cascade Prediction Loss}. The refinement targets are the same with the first stage yielding coarse prediction. 
%% 最后应该给出任务的总损失;
\begin{equation}
\mathcal{\hat{L}}^{\mathrm{OAD}}=-\sum_{i=t-m_S+1}^{t} \sum_{j=1}^{N_a} \boldsymbol{y}_{i, j} \log \hat{\boldsymbol{y}}_{i, j}
\end{equation}

\begin{equation}
\mathcal{\hat{L}}^{\mathrm{OAS}}=-\sum_{i=t-m_S-m_L+1}^{t-m_S} \sum_{j=1}^{N_a} \boldsymbol{y}_{i, j} \log \hat{\boldsymbol{y}}_{i, j}
\end{equation}

\noindent In summary, the training is guided by:
\begin{equation}
% \mathcal{L} =\lambda_1 \mathcal{L^{\mathrm{as}}} + \lambda_2 \mathcal{L^{\mathrm{oad}}}  
\mathcal{L} = \mathcal{L^{\mathrm{OAD}}} 
+ \lambda_1 \mathcal{L^{\mathrm{OAS}}} 
+ \lambda_2 \mathcal{\hat{L}^{\mathrm{OAD}}} 
+ \lambda_3 \mathcal{\hat{L}^{\mathrm{OAS}}} 
\end{equation}
where $\lambda_1$, $\lambda_2$, and $\lambda_3$ are three hyper-parameters to trade off different modules. 

% \begin{table}[t]
% \caption{Evaluation of the components of CWC-Trans on THUMOS'14.\label{tab:ablation}}
% \centering
% \begin{tabular}{ccc|c}
% \hline 
% Baseline & SWHD & WCR & mAP $(\%)$ \\
% \hline
% $\checkmark$ &  &  & $70.4$ \\
% $\checkmark$ & $\checkmark$ &  & $71.2$ \\
% $\checkmark$ & & $\checkmark$ & $70.8$ \\
% $\checkmark$ & \checkmark & $\checkmark$ & $71.4$ \\
% \hline 
% \multicolumn{3}{c|}{LSTR~\cite{xu2021long}} & $69.5$ \\
% \hline
% \end{tabular}
% \end{table}

\subsubsection{Online Inference}
Due to the characteristics of streaming data, it usually has to obtain the prediction results of each frame sequentially, as the well-known Streaming mode. In addition, LSTR~\cite{xu2021long} batches each frame by backtracking its corresponding short-term and long-term memory, which is named Batch mode. Similarly, we provide these two inference modes, where the Streaming mode is used for running time evaluation, and the other one is used for benchmarking performance. Note that both of them can produce the same performance. The diagram of streaming inference can be found in Figure~\ref{fig:teaser}.

\section{Experiments}
In this section, we will evaluate CWC-Trans on three public datasets, including THUMOS'14~\cite{idrees2017thumos}, TVSeries~\cite{geest2016online} and HDD~\cite{ramanishka2018CVPR}. We also provide detailed ablation studies on THUMOS'14~\cite{idrees2017thumos} to demonstrate the effectiveness of each component.

\subsection{Datasets}
\textbf{THUMOS'14}~\cite{idrees2017thumos} contains 200 and 213 untrimmed videos used for validation and test, respectively. These videos describe 20 classes of sports. On average, each video includes about 15.8 actions and 71\% background. Following the previous work~\cite{gao2017red}, we take the validation set for training and evaluate it on the test set.

\noindent\textbf{TVSeries}~\cite{geest2016online} has 27 untrimmed videos from 6 popular TV series, which is about 16 hours totally. The dataset is annotated with 30 realistic actions (e.g. run, eat, and \etc) at a frame level, and each action occurs at least 50 times. This dataset is very challenging because of the unconstrained perspectives, diversity of the actions, and a large proportion of background frames. 

\noindent\textbf{HDD}~\cite{ramanishka2018CVPR} includes 137 human driving sessions collected by vehicles equipped with various sensors, holding 104 hours. This dataset contains visual and various non-visual data and annotates 11 kinds of driving behaviours at a frame level. 
Following the previous work~\cite{wang2021oadtr}, we use vehicle sensor data as inputs, of which 100 sessions are used for training, and 37 sessions are used for testing.

\subsection{Implementation Details} 
\subsubsection{Feature Encoding}
Following the previous works~\cite{xu2021long,xu2019temporal,eun2020learning}, we extract video frames at 24 FPS. Then each second is split into four chunks so that each one has six frames. For feature encoding, we adopt TSN~\cite{wang2016temporal} pre-trained on Kinetics~\cite{carreira2017quo} and ActivityNet~\cite{caba2015activitynet}. The TSN is based on a two-stream network, where the spatial model and motion model adopt Resnet-50~\cite{he2016deep} and BN-Inception~\cite{ioffe2015batch}, respectively. To encode each chunk, the central frame (\ie, $4^{th}$) of each chunk and five consecutive optical flow maps computed by six frames are taken as inputs of the two streams, respectively. The visual and motion features are concatenated as the inputs of all the methods. 

\subsubsection{Hyperparameter Settings}
% 为了训练模型, 我们使用, 权重衰减XXX的Adam优化器来优化我们的模型。学习率首先线性增加至1e-4, 然后按照预先规律衰减到0。
% We train the model 25 epochs with batch size= 16; For XX module, we 把XX的值取为XXXX, XX
To train CWC-Trans, we use Adam~\cite{kingma2014adam} with a weight decay of $5 \times 10^{-5}$. The learning rate increases linearly to $1 \times 10^{-4}$ in the first 40\% of all the iterations. Then it decays to 0 according to the cosine function. We train the model by 25 epochs with a batch size of 16. As for hyper-parameters in CWC-Trans, we set the history length $m_L$ to 512, where one frame is sampled in each second. The trend length $m_S$ is 32, and the number of historical windows $N_{w}$ is 16. The dimension of history representations $d_L$ is 256, and the dimension of short-term trend $d_S$ is 1024. For the history encoder, we use 2 Swin stages and set the number of heads $N_h$ and $N_{h^{\prime}}$ to 4 and 8, respectively. The number of stages $M$ is 2. For the history decoder, the number of Swin layers $L$ at each stage is 4, 8, 4, and 2, respectively. The token expansion rate $r^{\prime}$ at each stage except the last stage is $\frac{1}{2}$, $\frac{1}{4}$, and $\frac{1}{4}$, respectively. For the current window encoder and cascade, the number of self-attention layers is 2. For the multi-task learning, $\lambda_1$, $\lambda_2$, and $\lambda_3$ are 0.2, 0.7, and 0.4, respectively.

\begin{table*}[ht]
\caption{\label{tab:sota}Online action detection performances on THUMOS'14, TVSeries and HDD. The mAP performance is reported for THUMOS'14 and HDD, while the mcAP performance is reported for TVSeries. The recent state-of-the-art methods on CNN, RNN, and Transformer are also included for comparison.}
\centering
\begin{tabular}{l|c|c|cc|cc|c}
\hline  
\multirow{2}{*}{Method} & \multirow{2}{*}{Venue}  & \multirow{2}{*}{Architecture} & \multicolumn{2}{c|}{THUMOS'14 / mAP (\%)} & \multicolumn{2}{c|}{TVSeries / mcAP (\%)} & \multicolumn{1}{c}{HDD / mAP (\%)} \\
\cline{4-8}& & & Kinetics & ActivityNet & Kinetics & ActivityNet & Sensor \\
\hline 
CDC~\cite{shou2017cdc}&CVPR 2017& CNN & - &44.4& - & - & -\\
RED~\cite{gao2017red} &CoRR 2017& RNN & - & 45.3& - & $79.2$ & $27.4$ \\ 
TRN~\cite{xu2019temporal} &ICCV 2019& RNN & $62.1$ &47.2 & $86.2$ & $83.7$ & 29.2 \\
IDN~\cite{eun2020learning} &CVPR 2020& RNN & $60.3$ & $50.0$ & $86.1$ & $84.7$ & -\\
LAP~\cite{qu2020lap}&CoRR 2020& RNN & - & $53.3$ & - & $85.3$ & - \\
FATS~\cite{YoungHwiKim2021TemporallySO}&PR 2021& RNN & $59.0$ & 51.6& $84.6$ & $81.7$ &-  \\
PKD~\cite{zhao2020privileged}&CVPR 2021& CNN & $64.5$ & - & $86.4$ & - &- \\
TFN~\cite{eun2021temporal}&PR 2021& CNN & - & $55.7$ & - & $85.0$ &- \\
WOAD~\cite{gao2021woad} &CVPR 2021& RNN & $67.1$ & - & - & - &- \\
OadTR~\cite{wang2021oadtr} &ICCV 2021& Transformer & $65.2$ & $58.3$ & $87.2$ & $85.4$ & $29.8$ \\
LSTR~\cite{xu2021long} &NeurIPS 2021& Transformer & $69.5$ & $65.3$ & $89.1$ & $88.1$ & - \\
GateHUB~\cite{chen2022gatehub}& CVPR 2022 & Transformer & $70.7$ & $69.1$ & $89.6$ & $88.4$ & $32.1$\\
\hline
CWC-Trans (Ours) &-& Transformer & \textbf{71.6} & \textbf{70.2} & \textbf{89.6} & \textbf{88.4} & \textbf{32.5} \\
\hline
\end{tabular}
\end {table*}

\begin{table}[t]
\caption{Evaluation of the components of CWC-Trans on THUMOS'14.\label{tab:ablation}}
\centering
\begin{tabular}{ccc|c}
\hline 
Baseline & SWHD & WCR & mAP $(\%)$ \\
\hline
$\checkmark$ &  &  & $70.4$ \\
$\checkmark$ & $\checkmark$ &  & $71.2$ \\
$\checkmark$ & & $\checkmark$ & $70.8$ \\
$\checkmark$ & \checkmark & $\checkmark$ & $71.4$ \\
\hline 
\multicolumn{3}{c|}{LSTR~\cite{xu2021long}} & $69.5$ \\
\hline
\end{tabular}
\end{table}

\begin{table*}[t]
    \caption{Circular window-based history encoder design. (a)~The impact of historical feature dimension. (b) The impact of different down-sampling strides. (c)~The impact of global self-attention on compression. (d)~The impact of different token reduction modules. (e)~The impact of different numbers of heads in MSTM.} 
    \label{tab:cwhe}
    \begin{subtable}{0.2\linewidth}
      \centering
      %\tiny
        \caption{\label{tab:cwhe:dim}}
        %\resizebox{!}{0.8cm}{
        \scalebox{0.9}{
        \begin{tabular}{l|c} 
            \hline 
            Method &  mAP $(\%)$ \\
            \hline 
            Dim=128 & $69.4$ \\
            Dim=256 & $70.4$ \\
            Dim=512 & $68.9$ \\
            Dim=1024 & $65.3$ \\
            \hline      
        \end{tabular}
        }
    \end{subtable}
    \begin{subtable}{0.19\linewidth}
        %\hspace{-5mm}
        \centering
        \caption{\label{tab:cwhe:ds}}
        %\resizebox{!}{0.8cm}{
        \scalebox{0.9}{
        \begin{tabular}{l|c} 
            \hline 
            Method &  mAP $(\%)$ \\
            \hline 
            % $\times 32, \times 4, \times 1$ & $69.7$\\
            % $\times 32, \times 8, \times 1$ & $70.2$\\
            % $\times 32, \times 8, \times 2, \times 1$ & $70.4$\\
            % $\times 32, \times 8, \times 4, \times 1$ & $69.6$\\
            $\times 8, \times 4 $ & $69.7$\\
            $\times 4, \times 8 $ & $70.2$\\
            $\times 4, \times 4, \times 2 $ & $70.4$\\
            $\times 4, \times 2, \times 4 $ & $69.6$\\
            \hline      
        \end{tabular}
        }
    \end{subtable} 
    \begin{subtable}{0.17\linewidth}
      \centering
        \caption{\label{tab:cwhe:gsa}}
        %\resizebox{!}{0.8cm}{
        \scalebox{0.9}{
        \begin{tabular}{l|c} 
            \hline 
            Method &  mAP $(\%)$ \\
            \hline 
            w/o SA & $70.1$ \\
            SA$\times$1 & $69.9$ \\
            SA$\times$2 & $69.4$ \\
            SA$\times$3 & $70.4$ \\
            \hline      
        \end{tabular}
        }
    \end{subtable} 
    %\hspace{-1mm}
    \begin{subtable}{0.2\linewidth}
      \centering
        \caption{\label{tab:cwhe:token_reduction}}
        %\resizebox{!}{0.8cm}{
        \scalebox{0.9}{
        \begin{tabular}{l|c} 
            \hline 
            Method &  mAP $(\%)$ \\
            \hline 
            PM & $69.5$ \\
            CA & $68.2$ \\
            TSM & $69.1$ \\ 
            MTSM (H=8) & $70.4$ \\
            \hline      
        \end{tabular}
        }
    \end{subtable}
    %\hspace{3.mm}
    \begin{subtable}{0.2\linewidth}
      \centering
        \caption{\label{tab:cwhe:mtsm}}
        %\resizebox{!}{0.8cm}{
        \scalebox{0.9}{
        \begin{tabular}{l|c} 
            \hline 
            Method &  mAP $(\%)$ \\
            \hline 
            MTSM (H = 2) & $70.0$ \\ 
            MTSM (H = 4) & $69.9$ \\
            MTSM (H = 8) & $70.4$ \\
            MTSM (H = 16) & $69.1$ \\
            \hline      
        \end{tabular}
        }
    \end{subtable} 
\end{table*}

\begin{table*}[t]
    \caption{Shifted window-based history decoder design.~(a) The impact of different MTSM structures.~(b) The impact of up-sampling strides.~(c) The impact of Swin layers at each stage.~(d) The impact of the shared action classifiers.}
    \label{tab:oas}
    \begin{subtable}[t]{0.3\linewidth}
      \centering
        \caption{\label{tab:oas:mtsm}}
        %\resizebox{!}{0.6cm}{
        \scalebox{0.9}{
        %\resizebox{\textwidth}{!}{
        \begin{tabular}{l|c} 
            \hline 
            Method &  mAP $(\%)$ \\
            \hline 
            MTSM (Local)& $70.6$ \\
            MTSM (Global)& $71.2$ \\
            MSTM (Global + Shortcut) & $ 70.4$ \\
            \hline      
        \end{tabular}
        }
    \end{subtable}
    %\hspace{5mm}
    \begin{subtable}[t]{0.22\linewidth}
      \centering
        \caption{\label{tab:oas:us}}
        %\resizebox{!}{0.6cm}{
        \scalebox{0.9}{
        \begin{tabular}{l|c} 
            \hline 
            Method &  mAP $(\%)$ \\
            \hline 
            % (4,8) 16,64,512 & $71.0$ \\
            % (2,16) 16,32,512 & $70.3$ \\
            $\times 2, \times 4, \times$ 4 & $71.2$ \\
            $\times 4, \times 2, \times$ 4 & $70.1$ \\
            $\times 4, \times 4, \times$ 2 & $70.5$ \\
            \hline      
        \end{tabular}
        }
    \end{subtable} 
    %\hspace{-5mm}
    \begin{subtable}[t]{0.2\linewidth}
      \centering
        \caption{\label{tab:oas:layer}}
        %\resizebox{!}{0.6cm}{
        \scalebox{0.9}{
        \begin{tabular}{l|c} 
            \hline 
            Method &  mAP $(\%)$ \\
            \hline 
            L = [2 8 4 4] & $70.4$ \\
            L = [4 8 4 4] & $70.9$ \\
            L = [4 6 4 4] & $70.2$ \\
            L = [4 8 4 2] & $71.2$ \\
            \hline      
        \end{tabular}
        }
    \end{subtable} 
    %\hspace{5mm}
    \begin{subtable}[t]{0.3\linewidth}
      \centering
      %\tiny
        \caption{\label{tab:oas:sc}}
        %\resizebox{!}{0.6cm}{
        \scalebox{0.9}{
        \begin{tabular}{l|c} 
            \hline 
            Method &  mAP $(\%)$ \\
            \hline 
            Shared classifier & $71.2$ \\
            Separate classifiers & $68.8$ \\
            \hline      
        \end{tabular}
        }
    \end{subtable} 
\end{table*}

\begin{table}[t]
\caption{Current window encoder design.~\label{tab:current}}
\centering
\begin{tabular}{l|c}
\hline 
Method &  mAP $(\%)$ \\
\hline 
w/ PE & $70.4$ \\
w/o PE & $67.9$ \\
w/o SA & $69.2$ \\
CA$\times$1 & $70.3$ \\
CA$\times$2 & $70.4$ \\
CA$\times$3 & $70.6$ \\
\hline
\end{tabular}
\end{table}

\subsection{Evaluation Metrics}
To evaluate the performance of online action detection, we adopt mean Average Precision (mAP) and mean calibrate average precision(mcAP)~\cite{geest2016online} in chunk-level, which means evaluation on 4 FPS, following the existing work~\cite{xu2021long,xu2019temporal,eun2020learning}. To calculate the mAP, we first need to calculate an average precision over all chunks for each action class and then average the average precision over all the action classes. mcAP~\cite{geest2016online} is proposed to correct the imbalance between background and actions interested.

\subsection{Comparison of the State-of-the-art Methods.} 
\subsubsection{Quantitative Results}
In this section, we compare the proposed CWC-Trans with the existing methods on three benchmark datasets to verify the effectiveness of our method, as shown in Table~\ref{tab:sota}. The architecture of these methods includes %3D convolution~\cite{shou2017cdc}, 
convolution~\cite{shou2017cdc,zhao2020privileged,eun2021temporal}, RNN~\cite{xu2019temporal,eun2020learning,qu2020lap,YoungHwiKim2021TemporallySO,gao2017red}, and Transformer~\cite{xu2021long,chen2022gatehub,wang2021oadtr}. Recently, Transformer-based methods have shown great advantages in this task. Our work designs an efficient window partitioning mechanism that can fully discover all the cues behind the long-term histories while ensuring a high inference speed. Following the previous work~\cite{xu2021long}, we take flow features and RGB features pre-trained on Kinetics~\cite{kay2017kinetics} or ActivityNet~\cite{caba2015activitynet} from TSN as the network inputs. From Table~\ref{tab:sota}, we
can observe that CWC-Trans outperforms all the existing works by 0.9\% and 1.1\% on THUMOS'14 when using features from Kinetics and ActivityNet, respectively. For TVSeries, CWC-Trans is slightly better than LSTR by 0.5\% and 0.3\% on features from Kinetics and ActivityNet, respectively, and is the same as GateHUB on both features. We believe that the metrics on TVSeries have been saturated, compared to other datasets. For HDD, CWC-Trans achieves state-of-the-art performance 32.5\% compared to GateHUB (32.1\%). 

\subsubsection{Qualitative Results}
\begin{figure*}
\centering
\subcaptionbox{Success cases on THUMUS (top two) and TVSeries (bottom two)\label{fig:vis:success}}{\includegraphics[width=15cm]{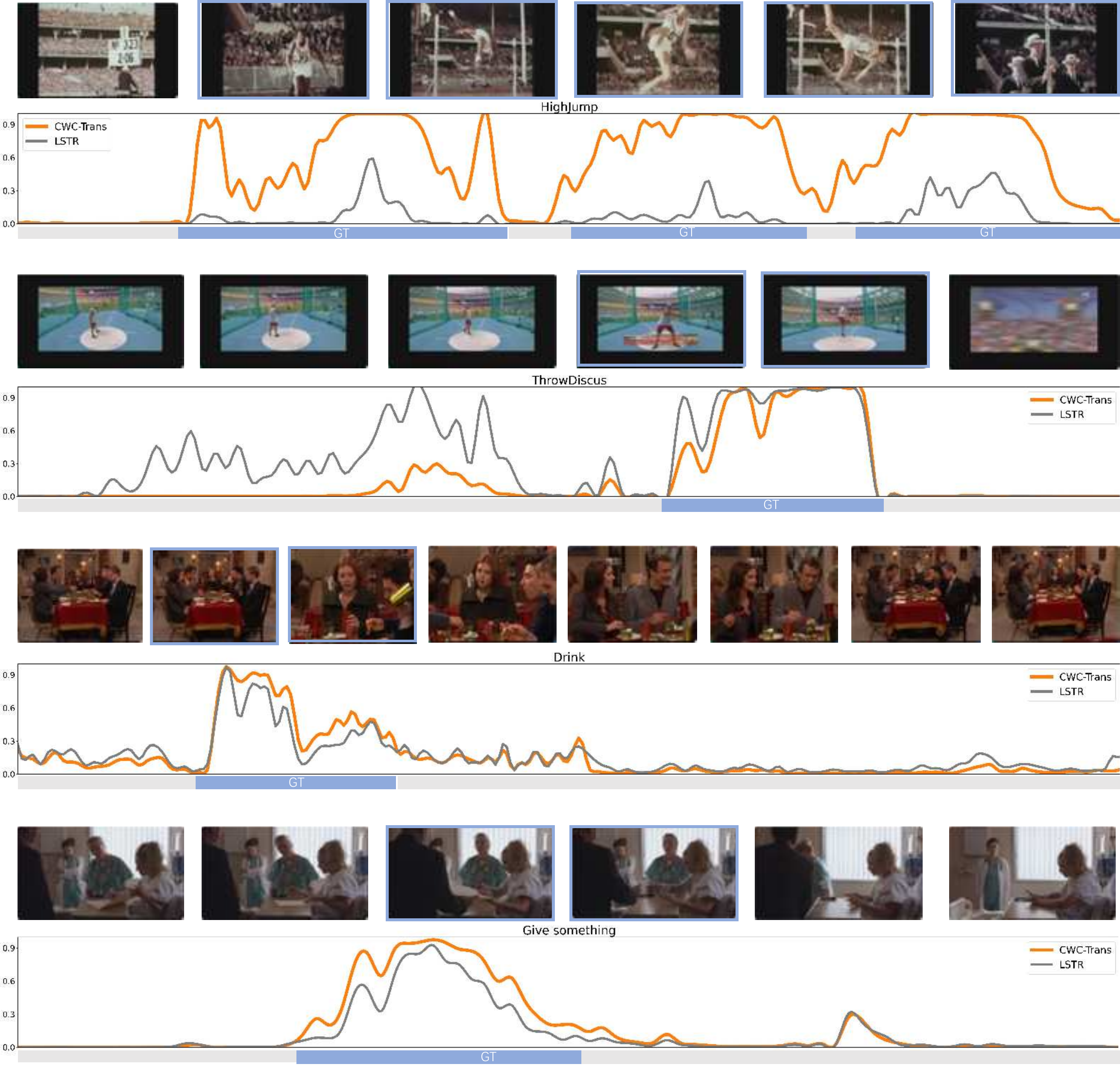}} \\
\subcaptionbox{Failure cases on THUMUS (top) and TVSeries (bottom)\label{fig:vis:failure}}{\includegraphics[width=15cm]{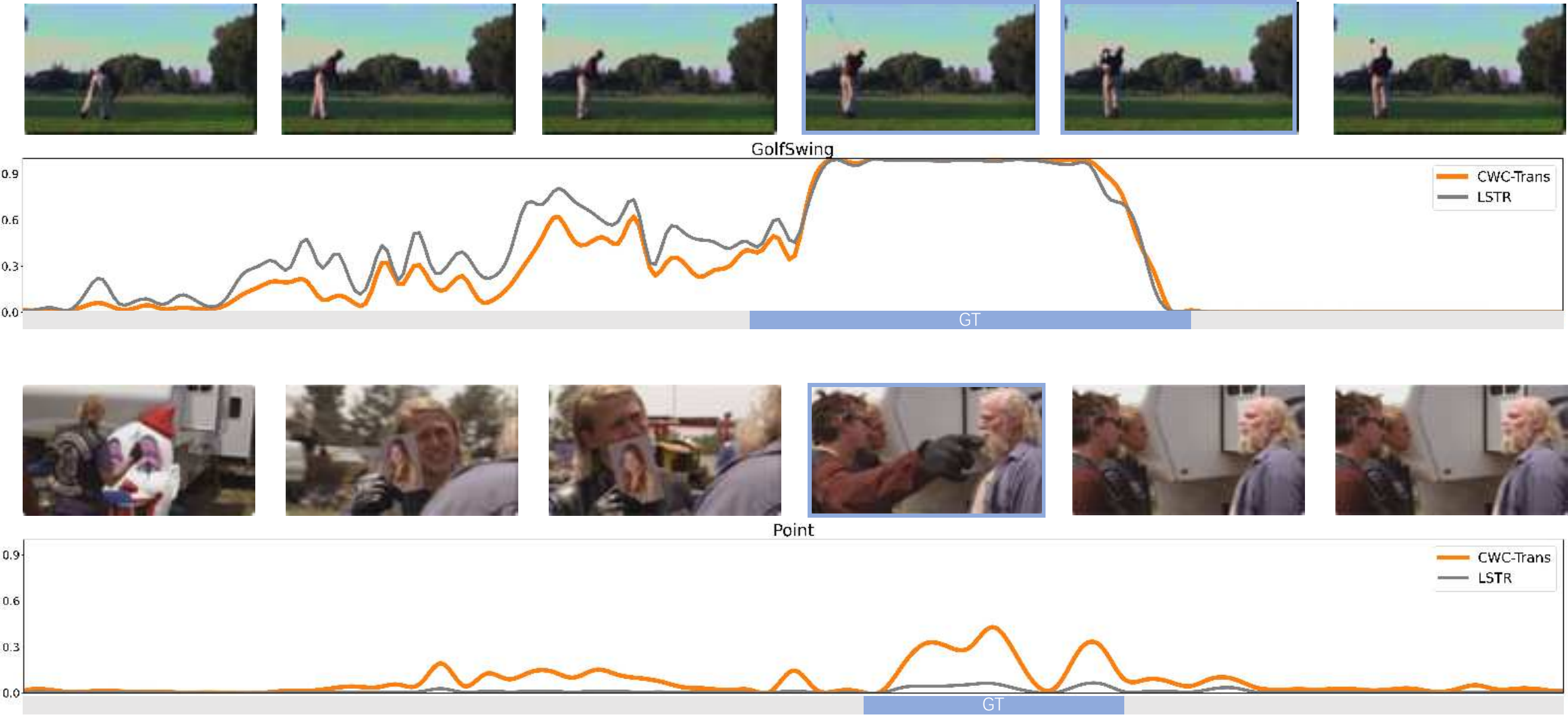}}
\caption{Visualization of success and failure cases of CWC-Trans and LSTR~\cite{xu2021long}.}
\end{figure*}

\textbf{Success Cases.} Figure~\ref{fig:vis:success} shows two success cases on THMOUS and TVSeries, respectively. Overall, our CWC-Trans outperforms LSTR on both datasets. Specifically, as shown in the first, third and fourth rows, our method is able to obtain higher prediction scores when the action occurs. This advantage is especially evident when the quick action, \eg, high jump, as shown in the first row. In addition, our method better suppresses motion interference thus has a lower false positive rate, as shown in the second row. These cases indicates that our hierarchical modeling strategy is able to obtain fine-grained temporal representations. 

\noindent\textbf{Failure Cases.} On the other hand, Figure~\ref{fig:vis:failure} shows two failure cases on THMOUS and TVSeries, respectively. 
As show in the first case, the ``GolfSwing`` action is quite similar to its preparatory movements, especially when the camera zooms out. Therefore, it is difficult to accurately distinguish the preparatory actions and the swing action, resulting in a large number of false positives. For the second case, the ``Point`` action is very abstract which usually happens suddenly and does not have enough contexts. Thus models usually fail to identify this action, though our proposed CWC-Trans achieves higher scores than LSTR. In summary, since the greater capability in modeling fine-grained temporal dependencies, our approach reports better results when dealing with these challenges.

% For TVSeries, we also evaluate CWC-Trans at different temporal ranges by a 10\% interval to see the impact of observation accumulation as time passes. \textcolor{blue}{As shown in Table~\ref{tab:tvportion}, }CWC-Trans also performs well at most temporal stages compared to LSTR and GateHUB.
% Besides, we also evaluate the performance at different action stages (10\% interval), as shown in Table~\ref{tab:ablation}. These results validate that our CWC-Trans consisting of \textcolor{red}{CWHE, OAS and cascade} preserves the critical clues, while compressing the long-term historical sequence to benefit the identification of the current moment. 

% \begin{table}[h!t]
% \caption{Performance of CWC-Trans on TVSeries and HDD.\label{tab:table2}}
% \centering
% \begin{tabular}{ccc}
% \hline
% \multirow{2}{*}{ Methods } & TVSeries & HDD \\
% & mcAP$(\%)$ & mAP$(\%)$ \\
% \hline 
% FATS~\cite{YoungHwiKim2021TemporallySO} & $84.6$ &  $-$ \\
% CNN~\cite{geest2016online} & $-$ &  $22.7$ \\
% LSTM~\cite{ramanishka2018CVPR} & $-$ &  $23.8$ \\
% IDN~\cite{eun2020learning} & $86.1$ &  $-$ \\
% RED~\cite{gao2017red} & $79.2$ &  $27.4$ \\
% TRN~\cite{xu2019temporal} & $86.2$ &  $29.2$ \\
% PDK~\cite{zhao2020privileged} & $86.4$ & $-$ \\
% OadTR~\cite{wang2021oadtr} & $87.2$ & $29.8$ \\
% LSTR~\cite{xu2021long}  & $89.1$ & $-$ \\
% GateHUB~\cite{chen2022gatehub}  & $89.6$ & $32.1$ \\
% \hline
% CWC-Trans(Ours) & $89.220$ & $-$ \\ 
% \hline 
% \end{tabular}
% \end{table}

\subsection{Ablation Study} 
% 可以对多个表格进行合并; 
In this section, we first conduct an ablation experiment to validate the effectiveness of each component of CWC-Trans, \ie, Circular Window-based History Encoder (CWHE), Shifted Window-based History Decoder (SWHD), and Window-based Cascade Refinement (WCR) for the OAD task. We then explore the impact of the hyper-parameters of each sub-module. Unless otherwise stated, all the experiments are conducted on THUMOS'14, using RGB and optical flow features from TSN. For simplicity, we report the mAP along with training for all the ablation studies. It is slightly different from online inference.

\subsubsection{Effectiveness of Each Component}
As shown in Table~\ref{tab:ablation}, we gradually append the proposed component to form four baselines as follows:

\noindent\textbf{Baseline.} Following our streaming video processing principle, we use the designed CWHE, which consists of window attention and MTSM, to compress the long-term historical sequence hierarchically. And then, the tokens from the current window cross-attend the compressed history for the final prediction of the current frame. 

\noindent\textbf{Baseline + SWHD.} We integrate the SWHD module with the baseline model to introduce the auxiliary task offline action segmentation. Besides, the logistic outputs of the SWHD are fed into the classifier that is shared with the outputs of the current window encoder.

\noindent\textbf{Baseline + WCR.} We only append the WCR module on the current window encoder to refine the baseline outputs. The final predictions are the outputs of the cascade module. 

\noindent\textbf{Baseline + SWHD + WCR (CWC-Trans).} It embodies all the carefully designed modules, i.e. CWHE, SWHD, and WCR. The WCR including another classifier that is also shared for OAS and OAD under this setting. 

The experimental results are summarized in Table.~\ref{tab:ablation}. We can observe that our \emph{Baseline} can achieve 70.4\%, which is better than LSTR by 0.9\%. It has already been demonstrated that the hierarchical window attention structure can compress historical sequences more effectively. Furthermore, with the help of the auxiliary task OAS, \emph{Baseline+SWHD} achieves a 0.8\% improvement over \emph{Baseline}. 
It indicates that the supervision signal from the OAS task can guide the compressed tokens to be more informative. Besides, \emph{Baseline+WCR} is 0.4\% higher than \emph{Baseline}, which indicates that the light-weight cascade module can lead to a progressive refinement on the coarse predictions.  \emph{Baseline+SWHD+WCR} achieves the best result over other baselines. 

% In addition to the advantages of SWHD and Cascade themselves, the shared classifier and Cascades modules between Current Window Encoder and SWHD strengthen the connection between the two tasks, i.e. OAS, OAD, thus significantly improving the performance. 

\subsubsection{Circular Window-based History Encoder Design} 

\begin{table*}[h!t]
    \caption{Cascade refinement design.~(a) The impact of different numbers of self-attention.~(b) The impact of the shared cascade and casual mask.~(c) The impact of different numbers of cascade stages.}
    \label{tab:cascade}
    \begin{subtable}[t]{0.33\linewidth}
      \centering
        \caption{\label{tab:cascade:sa}}
        %\resizebox{!}{0.8cm}{
        \scalebox{0.9}{
        \begin{tabular}{l|c} 
            \hline 
            Method &  mAP $(\%)$ \\
            \hline 
            SA$\times$2 w/o short cut & $66.2$ \\
            SA$\times$1 w/ short cut & $70.2$ \\
            SA$\times$2 w/ short cut & $70.8$ \\
            SA$\times$3 w/ short cut & $70.2$ \\
            \hline      
        \end{tabular}
        }
    \end{subtable}%
    \begin{subtable}[t]{0.33\linewidth}
      \centering
        \caption{\label{tab:cascade:share}}
        %\resizebox{!}{0.8cm}{
        \scalebox{0.9}{
        \begin{tabular}{l|c} 
            \hline 
            Method &  mAP $(\%)$ \\
            \hline 
            Shared cascade & $71.4$ \\ 
            Separate cascades &  $70.4$  \\
            OAS w/o casual mask & $65.5$ \\
            \hline      
        \end{tabular}
        }
    \end{subtable} 
    \begin{subtable}[t]{0.33\linewidth}
      \centering
        \caption{\label{tab:cascade:stage}}
        %\resizebox{!}{0.8cm}{
        \scalebox{0.9}{
        \begin{tabular}{l|c} 
            \hline 
            Method &  mAP $(\%)$ \\
            \hline 
            w/o cascade & $70.4$ \\
            cascade$\times$1 & $70.8$ \\
            cascade$\times$2 & $69.5$ \\
            \hline      
        \end{tabular}
        }
    \end{subtable} 
\end{table*}

% The proposed CWHE comprising of \textcolor{red}{ }. 
We conduct several experiments on \emph{Baseline} to explore the optimal network settings. All results are shown in Table~\ref{tab:cwhe}. 

%% 写法，简单介绍一下实验设置，然后列举结果, 然后根据结果得到结论; 

\noindent\textbf{History Feature Dimension Reduction.}
We project historical representations to a lower-dimensional feature space to ease the computational burden. Table~\ref{tab:cwhe:dim} varies feature dimension for initial history representations. The best performance is achieved when the dimension is 256, while inferior performance is observed when it is 128 or 512. But the performance drops dramatically when the dimension increases to 1024. We believe that much high feature dimension with a large number of historical frames makes optimization difficult.  

\noindent\textbf{Hierarchical Window Attention Design.} 
%% 需要补充消融实验，讲一下每个Stage所用的层数? 
%% 此外, 我们探索了Self-Attention的必要性; 
%% 窗口的尺寸对实验的影响, 我们把它去掉了。
As one of the core components of CWHE, at each stage, we perform window attention to model the dependencies between the tokens in each independent window. To reduce the number of parameters and mitigate over-fitting, the hierarchical attention module is shared across all historical windows. Each window will be finally squeezed into one representative vector $\boldsymbol{c}_i$. Here we study different down-sampling settings for window compression in Table~\ref{tab:cwhe:ds}, where the last down-sampling is mean pooling if the number of down-sampling is more than two. We observe gradually squeezing the windows is essential when performing hierarchical modeling, as classical network design~\cite{simonyan2014very, he2016deep, liu2021swin}.
Furthermore, hierarchical attention is performed separately in each window, so they lack global awareness of global history. Therefore, we add another attention module upon the window representations $\boldsymbol{C}$. As shown in Table~\ref{tab:cwhe:gsa}, it can be observed that when the sequence lacks global perception, \eg, \emph{w/o} SA, the performance drops by 0.3\%. As the number of SA layers increases, the performance gradually improves. We limit the number of global attention layers to three for inference efficiency. 

\noindent\textbf{Multi-head Token Slim Module.} 
The role of MTSM is to compress the number of tokens in each window at each stage. In Table~\ref{tab:cwhe:token_reduction}, we first compare MTSM with Patch Merging (PM), Cross-Attention (CA), and TSM. Our proposed MTSM has at least 0.9\% improvement and at most 2.2\% improvement. It can be explained PM merely fuses local Tokens without using global information for compression. As for CA, generic queries ignore the important data dependency for each window. However, TSM ignores the importance of different channels, though it realizes the data-dependent and global attention. Compared to TSM, our proposed MTSM improves the performance by 0.9\% even with only two heads, as shown in Table~\ref{tab:cwhe:mtsm}. The best 70.4\% mAP is achieved when the number of heads is set to 8.

% \begin{figure*}
% \centering
% \subcaptionbox{Success cases on THUMUS (top two) and TVSeries (bottom two)\label{fig:vis:success}}{\includegraphics[width=15cm]{figures/vis_success.pdf}} \\
% \subcaptionbox{Failure cases on THUMUS (top) and TVSeries (bottom)\label{fig:vis:failure}}{\includegraphics[width=15cm]{figures/vis_failure.pdf}}
% \caption{Visualization of success and failure cases of CWC-Trans and LSTR~\cite{xu2021long}.}
% \end{figure*}

\subsubsection{History Decoder Design} 
%% table
To perform the auxiliary task OAS, we design a decoder using Swin layers and MTSM to up-sample the compressed tokens. All the studies are conducted on the \emph{Baseline + SWHD}. As shown in Table~\ref{tab:oas}, we conduct extensive experiments on the structural design of the encoder, including how to use MTSM for upsampling, the number selection of Swin layers, and whether to share the classifier with the OAD task. For MTSM, MTSM (Local) indicates that each compressed token only needs to recover its single window without accessing other windows. MTSM (Global) indicates that MTSM takes all the compressed tokens as inputs, which means up-sampling from a global view. MTSM (Global + Shortcut) indicates that MTSM in the decoder needs to accept the outputs from the previous stage but also integrate the output with the same resolution from the encoder, which is an UNet~\cite{ronneberger2015u} style architecture. The comparison can be found in Table~\ref{tab:oas:mtsm}, the performance of MTSM (Global) is better than that of MTSM (Local). We believe the OAS supervision from a global view will provide holistic gradients for local window representation. However, we observe a significant performance drop after adding shortcuts in the global setting. We hypothesize that the outputs of the encoder will be attended for the OAD task while the shortcuts will weaken their representations, yielding a performance drop for OAD. Thus, we finally adopt the shortcut-free encoder-decoder for historical information modeling.
Additionally, we explored the token expansion rate at each stage in Table~\ref{tab:oas:us}. It turns out that the performance is best when the up-sampling rates are symmetric with the down-sampling rates in the encoder, \eg, $\times$4, $\times$4, and $\times$2. In Table~\ref{tab:oas:layer}, we also observe the optimal performance when the number of layers of each stage is separately set to 4, 8, 4, and 2. Surprisingly, in Table~\ref{tab:oas:sc} , we find that shared classifiers between OAS and OAD can significantly boost the performance by 2.4\%, compared to two separate classifiers. A reasonable explanation is that many historical tokens can be used as feature augmentation to train a comprehensive classifier, thus leading to a promising result for OAD.

\begin{table*}[t]
    \centering
    \caption{Efficiency comparison between our CWC-Trans and the previous work on parameter count (M) and inference speed (FPS). Our works outperforms GateHUB 0.9\% and 13.3\% on mAP and inference speed, respectively.}
    \label{tab:speed}
    \tabcolsep=2.5mm
\resizebox{\linewidth}{!}{
\begin{tabular}{l|c|cccc|c|c}
\hline \multirow{2}{*}{ Method} &  \multirow{2}{*}{Parameter Count}  & \multicolumn{5}{c|}{ Inference Speed (FPS) } & \multirow{2}{*}{$\mathrm{mAP}(\%)$} \\
\cline { 3 - 7 }& &  \begin{tabular}[c]{@{}l@{}}
Optical Flow \\ Computation
\end{tabular} & \begin{tabular}[c]{@{}l@{}}
RGB Feature \\ Extraction
\end{tabular}  & \begin{tabular}[c]{@{}l@{}}
Flow Feature \\ Extraction
\end{tabular}   & Model & Overall \\
\hline TRN  & $402.9 \mathrm{M}$ & $8.1$ & $70.5$ & $14.6$ & $123.3$ & $8.1$ & $62.1$ \\
OadTR & $75.8 \mathrm{M}$ & $8.1$ & $70.5$ & $14.6$ & $110.0$ & $8.1$ & $65.2$ \\
LSTR  & $58.0 \mathrm{M}$ & $8.1$ & $70.5$ & $14.6$ & $91.6$ & $8.1$ & $69.5$ \\
GateHUB & $45.2 \mathrm{M}$ & $8.1$ & $70.5$ & $14.6$ & $71.2$ & $8.1$ & ${70.7}$ \\
\hline 
CWC-Trans(Ours) & $56.7 \mathrm{M}$ & $8.1$ & $70.5$ & $14.6$ & $80.7$ & $8.1$ & $\mathbf{71.6}$ \\
\hline
\end{tabular}}
\end{table*}

\begin{figure}[t]
\centering
\includegraphics[height=5cm]{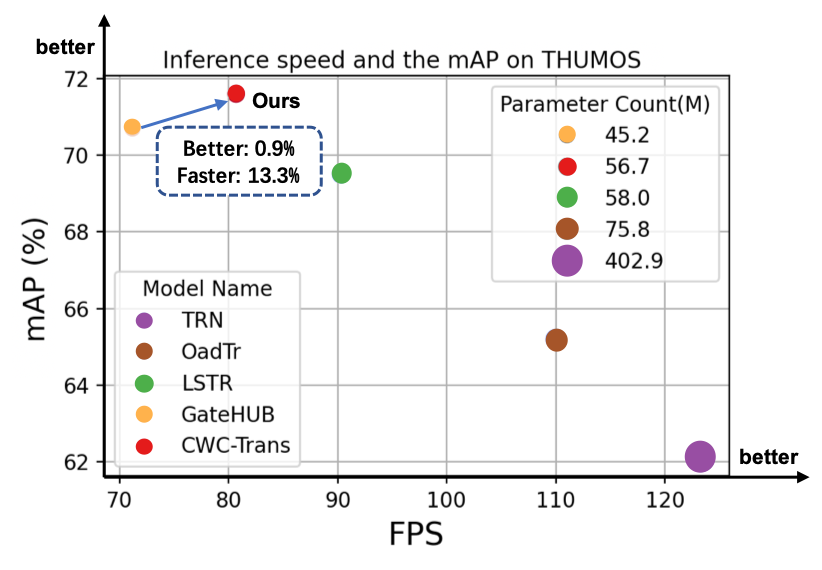}
\caption{Visualization of different methods on inference speed and the mAP
 on THUMOS'14.}
\label{fig:fps_map}
\end{figure}

\subsubsection{Current Window Encoder Design} 
We design several experiments to analyze the design of the current window encoder. All the studies are conducted on the \emph{Baseline}. Specifically, we experimentally analyze the impact of positional encoding(PE) and the number of the Cross-Attention (CA) modules. As shown in Table~\ref{tab:current}, \emph{w/o PE} indicates that positional encoding is not added to inputs of the current window. \emph{w/o SA} indicates that the Self-Attention (SA) is not incorporated before cross-attention. And CA $\times$ means the different numbers of the Cross-Attention (CA) module that includes one self-attention layer to fuse all the trend tokens and one cross-attention layer to fuse the history and trend.   
 From Table~\ref{tab:current}, we find that PE is vital for the current window inputs, and the performance drops by 2.5\% if PE is removed. Self-attention before cross attention is also helpful for OAD, where we observe a 1.2\% performance drop if self-attention is discarded. Besides, the performance is slightly improved when we increase the number of CA. Once again, we limit it to 2 for online inference. 
 
\subsubsection{Window-based Cascade Refinement Design}
 Here we study the design of the Window-based Cascade Refinement (WCR). Table~\ref{tab:cascade:sa} and Table~\ref{tab:cascade:stage} are conducted on the \emph{Baseline + WCR}, and Table~\ref{tab:cascade:share} is conducted on the \emph{CWC-Trans}. As shown in Table~\ref{tab:cascade:sa}, the mAP dramatically drops by 4.6\% if without shortcuts. This is because the probability inputs lose a lot of visual information. We believe that introducing shortcuts will significantly facilitate training the naive design of the WCR. In addition, we vary the number of SA at each stage and find that two SA layers are enough to achieve a good mAP. It also facilitates online inference due to the lightweight computations. Similar to the shared classifier, we also explore the effect of sharing WCR between OAS and OAD in Table~\ref{tab:cascade:share}. The shared WCR will bring 1\% improvement while reducing the parameters. Based on the shared WCR, if we remove the causal mask for OAS, each history token will attend to all. However, such attended history tokens will differ greatly from those in the current windows.
Consequently, the mAP (65.5\%) is much worse than the case (71.4\%) with the casual mask. At last, we explore the impact of the number of WCR, which can be observed from Table~\ref{tab:cascade:stage}. We find that even one WCR is enough to yield a better refinement on the coarse predictions. It will cause over-fitting (69.5\%) when the number of WCR increases to 3. This lightweight WCR enables us to achieve better performance without significant extra costs.

\subsection{Running Time}
%% 损失;
%% 网络的射界;
%% short cut; 
%% 是否利用History的分割概率?? 
%% 先不用写这个
We compare CWC-Trans with other methods in terms of running time, which is conducted on Tesla V100 with PyTorch 1.9. % The running time is averaged over all test videos in THUMOS dataset. 
We rerun LSTR on our platform since it is publicly available. Others are the reports from their manuscripts. In Table~\ref{tab:speed}, we report 80.7 FPS for CWC-Trans, which is 13.3\% faster and 0.9\% better than GateHUB, as illustrated in Figure~\ref{fig:fps_map}. Though CWC-Trans is slightly slower than LSTR, it has significantly better performance than LSTR (71.6\% versus 69.5\%). % We also observe bottlenecks for optical flow computation and feature extraction. One potential solution is to integrate these modules into an end-to-end network, which we leave in future work.  

\section{Conclusion}
In this paper, we carefully recap the existing methods of online action detection and point out some drawbacks, such as the absent feature pyramid and prediction refinement. We propose a novel window partition strategy called circular window with partial updating to address these issues. Based on it, we employ hierarchical modeling and cascade refinement on each window, which allows us to perform one single window updating during online inference. We also align online action detection with offline action segmentation, where we share the action classifier and the Window-based Cascade Refinement module to provide feature augmentation for online action detection. We provide extensive ablation studies on the evaluation of each component. Finally, the proposed CWC-Trans achieves state-of-the-art performance on THUMOS'14, TVSeries, and HDD, while it has 80.7 FPS running time. %For the OAD task, the existing setting is based on extracted features without spatial dimension. We believe that this prevents CWC-Trans from achieving optimal performance. One of our future research directions is to directly take image tokens rather than features as inputs, which pushes the OAD task forward fine-grained understanding. 

% \section{Acknowledge}

\bibliographystyle{IEEEtran}
\bibliography{reference}

\begin{thebibliography}{10}
\providecommand{\url}[1]{#1}
\csname url@samestyle\endcsname
\providecommand{\newblock}{\relax}
\providecommand{\bibinfo}[2]{#2}
\providecommand{\BIBentrySTDinterwordspacing}{\spaceskip=0pt\relax}
\providecommand{\BIBentryALTinterwordstretchfactor}{4}
\providecommand{\BIBentryALTinterwordspacing}{\spaceskip=\fontdimen2\font plus
\BIBentryALTinterwordstretchfactor\fontdimen3\font minus
  \fontdimen4\font\relax}
\providecommand{\BIBforeignlanguage}[2]{{%
\expandafter\ifx\csname l@#1\endcsname\relax
\typeout{** WARNING: IEEEtran.bst: No hyphenation pattern has been}%
\typeout{** loaded for the language `#1'. Using the pattern for}%
\typeout{** the default language instead.}%
\else
\language=\csname l@#1\endcsname
\fi
#2}}
\providecommand{\BIBdecl}{\relax}
\BIBdecl

\bibitem{hochreiter1997long}
S.~Hochreiter and J.~Schmidhuber, ``Long short-term memory,'' \emph{Neural
  computation}, vol.~9, no.~8, pp. 1735--1780, 1997.

\bibitem{cho2014learning}
K.~Cho, B.~Van~Merri{\"e}nboer, C.~Gulcehre, D.~Bahdanau, F.~Bougares,
  H.~Schwenk, and Y.~Bengio, ``Learning phrase representations using rnn
  encoder-decoder for statistical machine translation,'' \emph{arXiv preprint
  arXiv:1406.1078}, 2014.

\bibitem{wu2019long}
C.-Y. Wu, C.~Feichtenhofer, H.~Fan, K.~He, P.~Krahenbuhl, and R.~Girshick,
  ``Long-term feature banks for detailed video understanding,'' in
  \emph{Proceedings of the IEEE/CVF Conference on Computer Vision and Pattern
  Recognition}, 2019, pp. 284--293.

\bibitem{xu2021long}
M.~Xu, Y.~Xiong, H.~Chen, X.~Li, W.~Xia, Z.~Tu, and S.~Soatto, ``Long
  short-term transformer for online action detection,'' \emph{Advances in
  Neural Information Processing Systems}, vol.~34, 2021.

\bibitem{chen2022gatehub}
J.~Chen, G.~Mittal, Y.~Yu, Y.~Kong, and M.~Chen, ``Gatehub: Gated history unit
  with background suppression for online action detection,'' in
  \emph{Proceedings of the IEEE/CVF Conference on Computer Vision and Pattern
  Recognition}, 2022, pp. 19\,925--19\,934.

\bibitem{jaegle2021perceiver}
A.~Jaegle, F.~Gimeno, A.~Brock, O.~Vinyals, A.~Zisserman, and J.~Carreira,
  ``Perceiver: General perception with iterative attention,'' in
  \emph{International Conference on Machine Learning}.\hskip 1em plus 0.5em
  minus 0.4em\relax PMLR, 2021, pp. 4651--4664.

\bibitem{simonyan2014very}
K.~Simonyan and A.~Zisserman, ``Very deep convolutional networks for
  large-scale image recognition,'' \emph{arXiv preprint arXiv:1409.1556}, 2014.

\bibitem{he2016deep}
K.~He, X.~Zhang, S.~Ren, and J.~Sun, ``Deep residual learning for image
  recognition,'' in \emph{Proceedings of the IEEE conference on computer vision
  and pattern recognition}, 2016, pp. 770--778.

\bibitem{farha2019ms}
Y.~A. Farha and J.~Gall, ``Ms-tcn: Multi-stage temporal convolutional network
  for action segmentation,'' in \emph{Proceedings of the IEEE/CVF Conference on
  Computer Vision and Pattern Recognition}, 2019, pp. 3575--3584.

\bibitem{lin2017feature}
T.-Y. Lin, P.~Doll{\'a}r, R.~Girshick, K.~He, B.~Hariharan, and S.~Belongie,
  ``Feature pyramid networks for object detection,'' in \emph{Proceedings of
  the IEEE conference on computer vision and pattern recognition}, 2017, pp.
  2117--2125.

\bibitem{HaroonIdrees2017TheTC}
H.~Idrees, A.~R. Zamir, Y.-G. Jiang, A.~Gorban, I.~Laptev, R.~Sukthankar, and
  M.~Shah, ``The thumos challenge on action recognition for videos “in the
  wild”,'' \emph{Computer Vision and Image Understanding}, 2017.

\bibitem{geest2016online}
R.~D. Geest, E.~Gavves, A.~Ghodrati, Z.~Li, C.~Snoek, and T.~Tuytelaars,
  ``Online action detection,'' in \emph{European Conference on Computer
  Vision}.\hskip 1em plus 0.5em minus 0.4em\relax Springer, 2016, pp. 269--284.

\bibitem{ramanishka2018CVPR}
V.~Ramanishka, Y.-T. Chen, T.~Misu, and K.~Saenko, ``Toward driving scene
  understanding: A dataset for learning driver behavior and causal reasoning,''
  in \emph{Conference on Computer Vision and Pattern Recognition (CVPR)}, 2018.

\bibitem{de2018modeling}
R.~De~Geest and T.~Tuytelaars, ``Modeling temporal structure with lstm for
  online action detection,'' in \emph{2018 IEEE Winter Conference on
  Applications of Computer Vision (WACV)}.\hskip 1em plus 0.5em minus
  0.4em\relax IEEE, 2018, pp. 1549--1557.

\bibitem{simonyan2014two}
K.~Simonyan and A.~Zisserman, ``Two-stream convolutional networks for action
  recognition in videos,'' \emph{Advances in neural information processing
  systems}, vol.~27, 2014.

\bibitem{gao2017red}
J.~Gao, Z.~Yang, and R.~Nevatia, ``Red: Reinforced encoder-decoder networks for
  action anticipation,'' \emph{arXiv preprint arXiv:1707.04818}, 2017.

\bibitem{xu2019temporal}
M.~Xu, M.~Gao, Y.-T. Chen, L.~S. Davis, and D.~J. Crandall, ``Temporal
  recurrent networks for online action detection,'' in \emph{Proceedings of the
  IEEE/CVF International Conference on Computer Vision}, 2019, pp. 5532--5541.

\bibitem{wang2021oadtr}
X.~Wang, S.~Zhang, Z.~Qing, Y.~Shao, Z.~Zuo, C.~Gao, and N.~Sang, ``Oadtr:
  Online action detection with transformers,'' in \emph{Proceedings of the
  IEEE/CVF International Conference on Computer Vision}, 2021, pp. 7565--7575.

\bibitem{eun2020learning}
H.~Eun, J.~Moon, J.~Park, C.~Jung, and C.~Kim, ``Learning to discriminate
  information for online action detection,'' in \emph{Proceedings of the
  IEEE/CVF Conference on Computer Vision and Pattern Recognition}, 2020, pp.
  809--818.

\bibitem{qu2020lap}
S.~Qu, G.~Chen, D.~Xu, J.~Dong, F.~Lu, and A.~Knoll, ``Lap-net: Adaptive
  features sampling via learning action progression for online action
  detection,'' \emph{arXiv preprint arXiv:2011.07915}, 2020.

\bibitem{gao2021woad}
M.~Gao, Y.~Zhou, R.~Xu, R.~Socher, and C.~Xiong, ``Woad: Weakly supervised
  online action detection in untrimmed videos,'' in \emph{Proceedings of the
  IEEE/CVF Conference on Computer Vision and Pattern Recognition}, 2021, pp.
  1915--1923.

\bibitem{yang2022colar}
L.~Yang, J.~Han, and D.~Zhang, ``Colar: Effective and efficient online action
  detection by consulting exemplars,'' in \emph{Proceedings of the IEEE/CVF
  Conference on Computer Vision and Pattern Recognition}, 2022, pp. 3160--3169.

\bibitem{MarcusRohrbach2012ADF}
M.~Rohrbach, S.~Amin, M.~Andriluka, and B.~Schiele, ``A database for fine
  grained activity detection of cooking activities,'' \emph{computer vision and
  pattern recognition}, 2012.

\bibitem{karaman2014fast}
S.~Karaman, L.~Seidenari, and A.~Del~Bimbo, ``Fast saliency based pooling of
  fisher encoded dense trajectories,'' in \emph{ECCV THUMOS Workshop}, vol.~1,
  no.~2, 2014, p.~5.

\bibitem{HildeKuehne2015AnEG}
H.~Kuehne, J.~Gall, and T.~Serre, ``An end-to-end generative framework for
  video segmentation and recognition,'' \emph{workshop on applications of
  computer vision}, 2015.

\bibitem{KevinTang2012LearningLT}
K.~Tang, L.~Fei-Fei, and D.~Koller, ``Learning latent temporal structure for
  complex event detection,'' \emph{computer vision and pattern recognition},
  2012.

\bibitem{HildeKuehne2020AHR}
H.~Kuehne, A.~Richard, and J.~Gall, ``A hybrid rnn-hmm approach for weakly
  supervised temporal action segmentation,'' \emph{IEEE Transactions on Pattern
  Analysis and Machine Intelligence}, 2020.

\bibitem{BharatSingh2016AMB}
B.~Singh, T.~K. Marks, M.~Jones, O.~Tuzel, and M.~Shao, ``A multi-stream
  bi-directional recurrent neural network for fine-grained action detection,''
  \emph{computer vision and pattern recognition}, 2016.

\bibitem{ColinLea2016SegmentalSC}
C.~Lea, A.~Reiter, R.~Vidal, and G.~D. Hager, ``Segmental spatiotemporal cnns
  for fine-grained action segmentation,'' \emph{european conference on computer
  vision}, 2016.

\bibitem{AaronvandenOord2016WaveNetAG}
A.~van~den Oord, S.~Dieleman, H.~Zen, K.~Simonyan, O.~Vinyals, A.~Graves,
  N.~Kalchbrenner, A.~W. Senior, and K.~Kavukcuoglu, ``Wavenet: A generative
  model for raw audio,'' \emph{arXiv: Sound}, 2016.

\bibitem{ColinLea2016TemporalCN}
C.~Lea, M.~D. Flynn, R.~Vidal, A.~Reiter, and G.~D. Hager, ``Temporal
  convolutional networks for action segmentation and detection,''
  \emph{computer vision and pattern recognition}, 2016.

\bibitem{PengLei2018TemporalDR}
P.~Lei and S.~Todorovic, ``Temporal deformable residual networks for action
  segmentation in videos,'' \emph{computer vision and pattern recognition},
  2018.

\bibitem{FangqiuYi2022ASFormerTF}
F.~Yi, H.~Wen, and T.~Jiang, ``Asformer: Transformer for action segmentation,''
  2022.

\bibitem{du2022efficient}
D.~Du, B.~Su, Y.~Li, Z.~Qi, L.~Si, and Y.~Shan, ``Efficient u-transformer with
  boundary-aware loss for action segmentation,'' \emph{arXiv preprint
  arXiv:2205.13425}, 2022.

\bibitem{wang2022cross}
J.~Wang, Z.~Wang, S.~Zhuang, and H.~Wang, ``Cross-enhancement transformer for
  action segmentation,'' \emph{arXiv preprint arXiv:2205.09445}, 2022.

\bibitem{WeiLiu2016SSDSS}
W.~Liu, D.~Anguelov, D.~Erhan, C.~Szegedy, S.~Reed, C.-Y. Fu, and A.~C. Berg,
  ``Ssd: Single shot multibox detector,'' \emph{european conference on computer
  vision}, 2016.

\bibitem{TianweiLin2017SingleST}
T.~Lin, X.~Zhao, and Z.~Shou, ``Single shot temporal action detection,''
  \emph{acm multimedia}, 2017.

\bibitem{RunhaoZeng2020DenseRN}
R.~Zeng, H.~Xu, W.~Huang, P.~Chen, M.~Tan, and C.~Gan, ``Dense regression
  network for video grounding,'' \emph{computer vision and pattern
  recognition}, 2020.

\bibitem{ChumingLin2021LearningSB}
C.~Lin, C.~Xu, D.~Luo, Y.~Wang, Y.~Tai, C.~Wang, J.~Li, F.~Huang, and Y.~Fu,
  ``Learning salient boundary feature for anchor-free temporal action
  localization,'' \emph{computer vision and pattern recognition}, 2021.

\bibitem{zhang2022actionformer}
C.~Zhang, J.~Wu, and Y.~Li, ``Actionformer: Localizing moments of actions with
  transformers,'' \emph{arXiv preprint arXiv:2202.07925}, 2022.

\bibitem{ShyamalBuch2017SSTST}
S.~Buch, V.~Escorcia, C.~Shen, B.~Ghanem, and J.~C. Niebles, ``Sst:
  Single-stream temporal action proposals,'' \emph{computer vision and pattern
  recognition}, 2017.

\bibitem{TianweiLin2019BMNBN}
T.~Lin, X.~Liu, L.~Xin, E.~Ding, and S.~Wen, ``Bmn: Boundary-matching network
  for temporal action proposal generation,'' \emph{international conference on
  computer vision}, 2019.

\bibitem{RunhaoZeng2019GraphCN}
R.~Zeng, W.~Huang, M.~Tan, Y.~Rong, P.~Zhao, J.~Huang, and C.~Gan, ``Graph
  convolutional networks for temporal action localization,''
  \emph{international conference on computer vision}, 2019.

\bibitem{qing2021temporal}
Z.~Qing, H.~Su, W.~Gan, D.~Wang, W.~Wu, X.~Wang, Y.~Qiao, J.~Yan, C.~Gao, and
  N.~Sang, ``Temporal context aggregation network for temporal action proposal
  refinement,'' in \emph{Proceedings of the IEEE/CVF Conference on Computer
  Vision and Pattern Recognition}, 2021, pp. 485--494.

\bibitem{zhu2021enriching}
Z.~Zhu, W.~Tang, L.~Wang, N.~Zheng, and G.~Hua, ``Enriching local and global
  contexts for temporal action localization,'' in \emph{Proceedings of the
  IEEE/CVF International Conference on Computer Vision}, 2021, pp.
  13\,516--13\,525.

\bibitem{VictorEscorcia2016DAPsDA}
V.~Escorcia, F.~C. Heilbron, J.~C. Niebles, and B.~Ghanem, ``Daps: Deep action
  proposals for action understanding,'' \emph{european conference on computer
  vision}, 2016.

\bibitem{zhao2020bottom}
P.~Zhao, L.~Xie, C.~Ju, Y.~Zhang, Y.~Wang, and Q.~Tian, ``Bottom-up temporal
  action localization with mutual regularization,'' in \emph{European
  Conference on Computer Vision}.\hskip 1em plus 0.5em minus 0.4em\relax
  Springer, 2020, pp. 539--555.

\bibitem{YueranBai2020BoundaryCG}
Y.~Bai, Y.~Wang, Y.~Tong, Y.~Yang, Q.~Liu, and J.~Liu, ``Boundary content graph
  neural network for temporal action proposal generation,'' \emph{european
  conference on computer vision}, 2020.

\bibitem{JingTan2021RelaxedTD}
J.~Tan, J.~Tang, L.~Wang, and G.~Wu, ``Relaxed transformer decoders for direct
  action proposal generation,'' \emph{international conference on computer
  vision}, 2021.

\bibitem{YuanLiu2018MultigranularityGF}
Y.~Liu, L.~Ma, Y.~Zhang, W.~Liu, and S.-F. Chang, ``Multi-granularity generator
  for temporal action proposal,'' \emph{computer vision and pattern
  recognition}, 2018.

\bibitem{PeisenZhao2020BottomUpTA}
P.~Zhao, L.~Xie, C.~Ju, Y.~Zhang, Y.~Wang, and Q.~Tian, ``Bottom-up temporal
  action localization with mutual regularization,'' \emph{european conference
  on computer vision}, 2020.

\bibitem{ChenZhao2020VideoSG}
C.~Zhao, A.~Thabet, and B.~Ghanem, ``Video self-stitching graph network for
  temporal action localization,'' \emph{international conference on computer
  vision}, 2020.

\bibitem{MengmengXu2019GTADSL}
M.~Xu, C.~Zhao, D.~S. Rojas, A.~Thabet, and B.~Ghanem, ``G-tad: Sub-graph
  localization for temporal action detection,'' \emph{computer vision and
  pattern recognition}, 2019.

\bibitem{ShyamalBuch2017EndtoEndST}
S.~Buch, V.~Escorcia, B.~Ghanem, L.~Fei-Fei, and J.~C. Niebles, ``End-to-end,
  single-stream temporal action detection in untrimmed videos.'' \emph{british
  machine vision conference}, 2017.

\bibitem{LongFuchen2019GaussianTA}
L.~Fuchen, Y.~Ting, Q.~Zhaofan, T.~Xinmei, L.~Jiebo, and M.~Tao, ``Gaussian
  temporal awareness networks for action localization,'' \emph{IEEE Conference
  Proceedings}, 2019.

\bibitem{vaswani2017attention}
A.~Vaswani, N.~Shazeer, N.~Parmar, J.~Uszkoreit, L.~Jones, A.~N. Gomez,
  {\L}.~Kaiser, and I.~Polosukhin, ``Attention is all you need,''
  \emph{Advances in neural information processing systems}, vol.~30, 2017.

\bibitem{neimark2021video}
D.~Neimark, O.~Bar, M.~Zohar, and D.~Asselmann, ``Video transformer network,''
  in \emph{Proceedings of the IEEE/CVF International Conference on Computer
  Vision}, 2021, pp. 3163--3172.

\bibitem{sharir2021image}
G.~Sharir, A.~Noy, and L.~Zelnik-Manor, ``An image is worth 16x16 words, what
  is a video worth?'' \emph{arXiv preprint arXiv:2103.13915}, 2021.

\bibitem{li2021vidtr}
X.~Li, Y.~Zhang, C.~Liu, B.~Shuai, Y.~Zhu, B.~Brattoli, H.~Chen, I.~Marsic, and
  J.~Tighe, ``Vidtr: Video transformer without convolutions,'' \emph{arXiv
  e-prints}, pp. arXiv--2104, 2021.

\bibitem{bertasius2021space}
G.~Bertasius, H.~Wang, and L.~Torresani, ``Is space-time attention all you need
  for video understanding,'' \emph{arXiv preprint arXiv:2102.05095}, vol.~2,
  no.~3, p.~4, 2021.

\bibitem{arnab2021vivit}
A.~Arnab, M.~Dehghani, G.~Heigold, C.~Sun, M.~Lu{\v{c}}i{\'c}, and C.~Schmid,
  ``Vivit: A video vision transformer,'' in \emph{Proceedings of the IEEE/CVF
  International Conference on Computer Vision}, 2021, pp. 6836--6846.

\bibitem{nawhal2021activity}
M.~Nawhal and G.~Mori, ``Activity graph transformer for temporal action
  localization,'' \emph{arXiv preprint arXiv:2101.08540}, 2021.

\bibitem{tan2021relaxed}
J.~Tan, J.~Tang, L.~Wang, and G.~Wu, ``Relaxed transformer decoders for direct
  action proposal generation,'' in \emph{Proceedings of the IEEE/CVF
  International Conference on Computer Vision}, 2021, pp. 13\,526--13\,535.

\bibitem{dosovitskiy2020image}
A.~Dosovitskiy, L.~Beyer, A.~Kolesnikov, D.~Weissenborn, X.~Zhai,
  T.~Unterthiner, M.~Dehghani, M.~Minderer, G.~Heigold, S.~Gelly \emph{et~al.},
  ``An image is worth 16x16 words: Transformers for image recognition at
  scale,'' \emph{arXiv preprint arXiv:2010.11929}, 2020.

\bibitem{carion2020end}
N.~Carion, F.~Massa, G.~Synnaeve, N.~Usunier, A.~Kirillov, and S.~Zagoruyko,
  ``End-to-end object detection with transformers,'' in \emph{European
  conference on computer vision}.\hskip 1em plus 0.5em minus 0.4em\relax
  Springer, 2020, pp. 213--229.

\bibitem{parmar2018image}
N.~Parmar, A.~Vaswani, J.~Uszkoreit, L.~Kaiser, N.~Shazeer, A.~Ku, and D.~Tran,
  ``Image transformer,'' in \emph{International conference on machine
  learning}.\hskip 1em plus 0.5em minus 0.4em\relax PMLR, 2018, pp. 4055--4064.

\bibitem{child2019generating}
R.~Child, S.~Gray, A.~Radford, and I.~Sutskever, ``Generating long sequences
  with sparse transformers,'' \emph{arXiv preprint arXiv:1904.10509}, 2019.

\bibitem{ho2019axial}
J.~Ho, N.~Kalchbrenner, D.~Weissenborn, and T.~Salimans, ``Axial attention in
  multidimensional transformers,'' \emph{arXiv preprint arXiv:1912.12180},
  2019.

\bibitem{liu2021swin}
Z.~Liu, Y.~Lin, Y.~Cao, H.~Hu, Y.~Wei, Z.~Zhang, S.~Lin, and B.~Guo, ``Swin
  transformer: Hierarchical vision transformer using shifted windows,'' in
  \emph{Proceedings of the IEEE/CVF International Conference on Computer
  Vision}, 2021, pp. 10\,012--10\,022.

\bibitem{wang2020linformer}
S.~Wang, B.~Z. Li, M.~Khabsa, H.~Fang, and H.~Ma, ``Linformer: Self-attention
  with linear complexity,'' \emph{arXiv preprint arXiv:2006.04768}, 2020.

\bibitem{xiong2021nystromformer}
Y.~Xiong, Z.~Zeng, R.~Chakraborty, M.~Tan, G.~Fung, Y.~Li, and V.~Singh,
  ``Nystr{\"o}mformer: A nystr{\"o}m-based algorithm for approximating
  self-attention,'' in \emph{Proceedings of the AAAI Conference on Artificial
  Intelligence}, vol.~35, no.~16, 2021, pp. 14\,138--14\,148.

\bibitem{tay2021synthesizer}
Y.~Tay, D.~Bahri, D.~Metzler, D.-C. Juan, Z.~Zhao, and C.~Zheng, ``Synthesizer:
  Rethinking self-attention for transformer models,'' in \emph{International
  conference on machine learning}.\hskip 1em plus 0.5em minus 0.4em\relax PMLR,
  2021, pp. 10\,183--10\,192.

\bibitem{rae2019compressive}
J.~W. Rae, A.~Potapenko, S.~M. Jayakumar, and T.~P. Lillicrap, ``Compressive
  transformers for long-range sequence modelling,'' \emph{arXiv preprint
  arXiv:1911.05507}, 2019.

\bibitem{lee2019set}
J.~Lee, Y.~Lee, J.~Kim, A.~Kosiorek, S.~Choi, and Y.~W. Teh, ``Set transformer:
  A framework for attention-based permutation-invariant neural networks,'' in
  \emph{International conference on machine learning}.\hskip 1em plus 0.5em
  minus 0.4em\relax PMLR, 2019, pp. 3744--3753.

\bibitem{choromanski2020rethinking}
K.~Choromanski, V.~Likhosherstov, D.~Dohan, X.~Song, A.~Gane, T.~Sarlos,
  P.~Hawkins, J.~Davis, A.~Mohiuddin, L.~Kaiser \emph{et~al.}, ``Rethinking
  attention with performers,'' \emph{arXiv preprint arXiv:2009.14794}, 2020.

\bibitem{katharopoulos2020transformers}
A.~Katharopoulos, A.~Vyas, N.~Pappas, and F.~Fleuret, ``Transformers are rnns:
  Fast autoregressive transformers with linear attention,'' in
  \emph{International Conference on Machine Learning}.\hskip 1em plus 0.5em
  minus 0.4em\relax PMLR, 2020, pp. 5156--5165.

\bibitem{peng2021random}
H.~Peng, N.~Pappas, D.~Yogatama, R.~Schwartz, N.~A. Smith, and L.~Kong,
  ``Random feature attention,'' \emph{arXiv preprint arXiv:2103.02143}, 2021.

\bibitem{feichtenhofer2019slowfast}
C.~Feichtenhofer, H.~Fan, J.~Malik, and K.~He, ``Slowfast networks for video
  recognition,'' in \emph{Proceedings of the IEEE/CVF international conference
  on computer vision}, 2019, pp. 6202--6211.

\bibitem{arandjelovic2016netvlad}
R.~Arandjelovic, P.~Gronat, A.~Torii, T.~Pajdla, and J.~Sivic, ``Netvlad: Cnn
  architecture for weakly supervised place recognition,'' in \emph{Proceedings
  of the IEEE conference on computer vision and pattern recognition}, 2016, pp.
  5297--5307.

\bibitem{zong2021self}
Z.~Zong, K.~Li, G.~Song, Y.~Wang, Y.~Qiao, B.~Leng, and Y.~Liu, ``Self-slimmed
  vision transformer,'' \emph{arXiv preprint arXiv:2111.12624}, 2021.

\bibitem{idrees2017thumos}
H.~Idrees, A.~R. Zamir, Y.-G. Jiang, A.~Gorban, I.~Laptev, R.~Sukthankar, and
  M.~Shah, ``The thumos challenge on action recognition for videos “in the
  wild”,'' \emph{Computer Vision and Image Understanding}, vol. 155, pp.
  1--23, 2017.

\bibitem{wang2016temporal}
L.~Wang, Y.~Xiong, Z.~Wang, Y.~Qiao, D.~Lin, X.~Tang, and L.~V. Gool,
  ``Temporal segment networks: Towards good practices for deep action
  recognition,'' in \emph{European conference on computer vision}.\hskip 1em
  plus 0.5em minus 0.4em\relax Springer, 2016, pp. 20--36.

\bibitem{carreira2017quo}
J.~Carreira and A.~Zisserman, ``Quo vadis, action recognition? a new model and
  the kinetics dataset,'' in \emph{proceedings of the IEEE Conference on
  Computer Vision and Pattern Recognition}, 2017, pp. 6299--6308.

\bibitem{caba2015activitynet}
F.~Caba~Heilbron, V.~Escorcia, B.~Ghanem, and J.~Carlos~Niebles, ``Activitynet:
  A large-scale video benchmark for human activity understanding,'' in
  \emph{Proceedings of the ieee conference on computer vision and pattern
  recognition}, 2015, pp. 961--970.

\bibitem{ioffe2015batch}
S.~Ioffe and C.~Szegedy, ``Batch normalization: Accelerating deep network
  training by reducing internal covariate shift,'' in \emph{International
  conference on machine learning}.\hskip 1em plus 0.5em minus 0.4em\relax PMLR,
  2015, pp. 448--456.

\bibitem{kingma2014adam}
D.~P. Kingma and J.~Ba, ``Adam: A method for stochastic optimization,''
  \emph{arXiv preprint arXiv:1412.6980}, 2014.

\bibitem{shou2017cdc}
Z.~Shou, J.~Chan, A.~Zareian, K.~Miyazawa, and S.-F. Chang, ``Cdc:
  Convolutional-de-convolutional networks for precise temporal action
  localization in untrimmed videos,'' in \emph{Proceedings of the IEEE
  conference on computer vision and pattern recognition}, 2017, pp. 5734--5743.

\bibitem{YoungHwiKim2021TemporallySO}
Y.~H. Kim, S.~Nam, and S.~J. Kim, ``Temporally smooth online action detection
  using cycle-consistent future anticipation,'' \emph{Pattern Recognition},
  2021.

\bibitem{zhao2020privileged}
P.~Zhao, L.~Xie, Y.~Zhang, Y.~Wang, and Q.~Tian, ``Privileged knowledge
  distillation for online action detection,'' \emph{arXiv preprint
  arXiv:2011.09158}, 2020.

\bibitem{eun2021temporal}
H.~Eun, J.~Moon, J.~Park, C.~Jung, and C.~Kim, ``Temporal filtering networks
  for online action detection,'' \emph{Pattern Recognition}, vol. 111, p.
  107695, 2021.

\bibitem{kay2017kinetics}
W.~Kay, J.~Carreira, K.~Simonyan, B.~Zhang, C.~Hillier, S.~Vijayanarasimhan,
  F.~Viola, T.~Green, T.~Back, P.~Natsev \emph{et~al.}, ``The kinetics human
  action video dataset,'' \emph{arXiv preprint arXiv:1705.06950}, 2017.

\bibitem{ronneberger2015u}
O.~Ronneberger, P.~Fischer, and T.~Brox, ``U-net: Convolutional networks for
  biomedical image segmentation,'' in \emph{International Conference on Medical
  image computing and computer-assisted intervention}.\hskip 1em plus 0.5em
  minus 0.4em\relax Springer, 2015, pp. 234--241.

\end{thebibliography}

\end{document}